\documentclass{article}

\usepackage{PRIMEarxiv}

\usepackage[utf8]{inputenc} % allow utf-8 input
\usepackage[T1]{fontenc}    % use 8-bit T1 fonts
\usepackage{hyperref}       % hyperlinks
\usepackage{url}            % simple URL typesetting
\usepackage{booktabs}       % professional-quality tables
\usepackage{amsfonts}       % blackboard math symbols
\usepackage{nicefrac}       % compact symbols for 1/2, etc.
\usepackage{microtype}      % microtypography
\usepackage{lipsum}
\usepackage{fancyhdr}       % header
\usepackage{graphicx}       % graphics
\graphicspath{{media/}}     % organize your images and other figures under media/ folder
\usepackage[version=4]{mhchem}
\usepackage[font=small,labelfont=bf]{caption}
\usepackage{float}
\usepackage{amsthm}

\usepackage[utf8]{inputenc} 
\usepackage{geometry} 
\usepackage{tikz} 
\usepackage{lipsum}

\title{\bf Higher-Order Knowledge Representations for Agentic Scientific Reasoning}

\author{
    Isabella A. Stewart \\
    Department of Civil and Environmental Engineering \\
    Massachusetts Institute of Technology \\
    Cambridge, MA, USA
\And
    Markus J. Buehler \\
    Department of Civil and Environmental Engineering \\
    Department of Mechanical Engineering \\
    Schwarzman College of Computing \\
    Massachusetts Institute of Technology \\
    Cambridge, MA, USA \\
    \\
    Corresponding author: \texttt{mbuehler@mit.edu}
}

\begin{document}
\raggedbottom
\maketitle

\begin{abstract}

Scientific inquiry requires systems-level reasoning that integrates heterogeneous experimental data, cross-domain knowledge, and mechanistic evidence into coherent explanations. While Large Language Models (LLMs) offer inferential capabilities, they often depend on retrieval-augmented contexts that lack structural depth. Traditional Knowledge Graphs (KGs) attempt to bridge this gap, yet their pairwise constraints fail to capture the irreducible higher-order interactions that govern emergent physical behavior. To address this, we introduce a methodology for constructing hypergraph-based knowledge representations that faithfully encode multi-entity relationships. Applied to a corpus of $\approx$ 1,100 manuscripts on biocomposite scaffolds, our framework constructs a global hypergraph of 161,172 nodes and 320,201 hyperedges, revealing a scale-free topology (power law exponent $\approx$ 1.23) organized around highly connected conceptual hubs. This representation prevents the combinatorial explosion typical of pairwise expansions and explicitly preserves the co-occurrence context of scientific formulations. We further demonstrate that equipping agentic systems with hypergraph traversal tools, specifically using node-intersection constraints, enables them to bridge semantically distant concepts. By exploiting these higher-order pathways, the system successfully generates grounded mechanistic hypotheses for novel composite materials, such as linking cerium oxide to PCL scaffolds via chitosan intermediates. This work establishes a ``teacherless'' agentic reasoning system where hypergraph topology acts as a verifiable guardrail, accelerating scientific discovery by uncovering relationships obscured by traditional graph methods.

\color{black}
\end{abstract}


\section{Introduction}
Large language models (LLMs) have demonstrated capabilities in natural language processing and generation \cite{vaswani2023attentionneed, JMLR:v21:20-074, devlin-etal-2019-bert, chowdhery2022palmscalinglanguagemodeling}. However, these models fundamentally encode knowledge through implicit parametric representations distributed across learned weights, rendering factual information difficult to access, verify, or systematically update \cite{petroni-etal-2019-language, roberts-etal-2020-much, meng2023locatingeditingfactualassociations, 10.1145/3627673.3679673, gekhman2025insideouthiddenfactualknowledge, mousavi2025llmsrepositoriesfactualknowledge, mallen2023trustlanguagemodelsinvestigating, modarressi2025memllmfinetuningllmsuse, thede2025wikibigeditunderstandinglimitslifelong,Buehler2024}. This architectural constraint manifests in well-documented failure modes including factual hallucination, temporal knowledge degradation, and diminished performance on long-tail or domain-specific queries \cite{10.1145/3703155, maynez-etal-2020-faithfulness, brown2020languagemodelsfewshotlearners, jiang2020knowlanguagemodelsknow}. 

Prior work has established that generative AI can drive the \textit{de novo }design of complex material architectures from natural language prompts, and that multi-agent systems can automate the discovery of a range of materials and systems including engineering, protein and alloy design principles by integrating simulation tools with agentic reasoning engines~\cite{D4DD00013G,Lu2025FineTuningLLMs,ghafarollahi2025sparksmultiagentartificialintelligence,RN31,Ghafarollahi_Buehler_2025,Buehler2025GPReFLexOR}. Building on these foundations, recent work has explored hybrid approaches that augment LLMs with structured, non-parametric knowledge representations such as knowledge graphs, formal ontologies, and relational databases through in-context learning (ICL) mechanisms \cite{Hu2023,Buehler2024, Ghafarollahi2024, jiang-etal-2023-structgpt, hao2022structuredpromptingscalingincontext, lee2024learningreduceoptimalrepresentations, waghmare2025efficientgraphunderstandingllms}. Other work incorporated graph-reasoning capabilities directly into the Transformer architecture, as reported in~\cite{Buehler2025GA}, and a range of research has utilized graph neural networks (GNNs) in describing a range of natural phenomena~\cite{Reiser2022,Zhou2020,Yang2025MultiCell}. 

In contrast to retrieval-augmented generation methods that primarily inject unstructured text, structured representations provide explicit relational semantics that guide reasoning processes. When appropriately serialized within prompts, these representations enable LLMs to execute complex tasks without parameter updates or task-specific fine-tuning. From a practical standpoint, this allows knowledge to be updated by modifying external structures rather than retraining the model, which is essential in fast-moving fields like biomedicine or materials science. With this approach, domain-specific knowledge graphs can rapidly specialize general-purpose LLMs \cite{li_are_2025, hu_lets_2024, wang_enhancing_2025}. External knowledge sources also provide explicit provenance chains, enabling output auditing and error attribution essential for high-stakes applications in healthcare, legal reasoning, and scientific research, with some successes \cite{matsumoto_kragen_2024, almuntashiri_using_2025}. 

Recent work has shown that augmenting LLMs with graph-based and ontological knowledge representations facilitates compositional reasoning across disparate knowledge domains and the discovery of previously unseen relationships \cite{Buehler2024, Ghafarollahi2024, pan_unifying_2023, sharma_og-rag_2024}. In such frameworks, generative AI can facilitate the construction of bridges across heterogeneous knowledge domains by generating analogies, proposing novel associations, and offering explanatory connections between concepts that may initially appear unrelated. With this structured knowledge graph in context for ICL, the model can precisely delineate between interconceptual relationships and engage in a form of reasoning that parallels human creative and scientific thought \cite{sharma_og-rag_2024}. Fundamentally, innovation, discovery, and creative cognition depend on the capacity to traverse intricate conceptual landscapes -- whether to formulate hypotheses, explain emergent phenomena, or predict the behavior of previously unstudied systems. These cognitive processes can be understood as a form of pathfinding through a latent space wherein coherent trajectories are constructed to yield novel insights.

In practice, knowledge is often encoded into graph networks by representing concepts as nodes and their relationships as edges, which capture pairwise correlations between structures \cite{Ehrlinger2016TowardsAD, Hogan_2021, Buehler2024, Liu2024}. Each edge reflects the direct connection or relationship between two specific concepts, allowing the network to represent how individual ideas are related to one another \cite{Ehrlinger2016TowardsAD, Hogan_2021}. This pairwise structure encodes meaningful relationships, such as causality, similarity, or other types of connections, forming pathways for reasoning. Through techniques like data mining and embedding models, these graphs become enriched with semantic depth \cite{grover2016node2vecscalablefeaturelearning, Tang_2015}. Embeddings capture latent similarities between concepts, even when explicit edges are absent, enabling the system to propose plausible but previously unrecognized connections \cite{Buehler2024, Ghafarollahi2024, Buehler2025GPReFLexOR, Buehler2025SelfOrganizingGraphReasoning}.

We posit here that traditional pairwise KGs are, however, ill-suited for scientific reasoning as they cannot adequately capture higher-order interactions among multiple entities that often govern emergent physical system behavior. We therefore introduce a methodology for constructing hypergraph-based knowledge representations that move beyond traditional pairwise graphs. 

\subsection{Standard Graph Preliminary: Representing Pairwise Relations in Graph Structure}

Formally, standard graph theory can be defined with the following preliminary. A Knowledge Graph (KG) $\mathcal{G}(\mathcal{E}, \mathcal{R}, \mathcal{T})$ consists of a set of entities $\mathcal{E}$, relations $\mathcal{R}$, and knowledge triples $\mathcal{T} \subseteq \mathcal{E} \times \mathcal{R} \times \mathcal{E}$. Each triple $T = (e_h, r, e_t)$ denotes a factual edge in $\mathcal{G}$. For a subset $\mathcal{E}_S \subseteq \mathcal{E}$, the induced subgraph is $\mathcal{S} = (\mathcal{E}_S, \mathcal{R}_S, \mathcal{T}_S)$ with:

\begin{itemize}
    \item $\mathcal{T}_S = \{(e, r, e') \in \mathcal{T} \mid e, e' \in \mathcal{E}_S\}$
    \item $\mathcal{R}_S = \{r \in \mathcal{R} \mid (e, r, e') \in \mathcal{T}_S\}$
\end{itemize}

Let $\mathcal{D}(e)$ and $\mathcal{D}(r)$ denote the textual descriptions of entity $e \in \mathcal{E}$ and relation $r \in \mathcal{R}$.

\paragraph{Definition 1 (Reasoning Path).} 
A reasoning path from $e_1$ to $e_{l+1}$ is a sequence of triples with length $l$:
\[
\text{path}_{\mathcal{G}}(e_1, e_{l+1}) = \{(e_1, r_1, e_2), (e_2, r_2, e_3), \dots, (e_l, r_l, e_{l+1})\}
\]

\medskip
\noindent\textit{Example.}
A reasoning path from “Sally” to “Bob” might be:
\[
(\text{Sally}, \text{is co-authors with}, \text{David}) \rightarrow (\text{David}, \text{is co-authors with}, \text{Ella}) \rightarrow (\text{Ella}, \text{is co-authors with}, \text{Bob})
\]
This path has length 3.

\medskip
\noindent\textit{Distance and Neighborhood.}
If a reasoning path exists between $s$ and $t$, we write $s \leftrightarrow t$. The distance between $s$ and $t$ in $\mathcal{G}$ is defined as the length of the shortest such path, denoted $\text{dist}_{\mathcal{G}}(s, t)$. If no path exists, $\text{dist}_{\mathcal{G}}(s, t) = \infty$. The $h$-hop neighborhood of $s$ is given by:
\[
N_{\mathcal{G}}(s, h) = \{t \in \mathcal{E} \mid \text{dist}_{\mathcal{G}}(s, t) \leq h\}
\]

\paragraph{Definition 2 (Entity Path).}
For a list $\text{list}_e = [e_1, e_2, \dots, e_l]$, the entity path is the union of reasoning paths between each consecutive pair:
\[
\text{path}_{\mathcal{G}}(\text{list}_e) = \bigcup_{1 \leq i < l} \text{path}_{\mathcal{G}}(e_i, e_{i+1})
\]

\subsection{Extending Pairwise Graphs to Higher-Order Network Models}
In many real-world applications, relationships among entities extend beyond simple pairwise interactions \cite{bick_what_2023, bretto_hypergraph_2013, chodrow_configuration_2020}. Consider chemical reactions: a single reaction often involves multiple reactants and products interacting simultaneously, not just in isolated pairs. Naturally occurring topologies further illustrate this higher-order organization: in vascular plants, early species exhibit tree-like branching structures that expose their distribution system to single-point failure, whereas more recent species evolve dense, nested loop architectures that enhance resilience, and analogously, neuronal networks can sustain complex firing behavior, where a single firing event in an upstream neuron triggers persistent, multi-neuron activation through embedded feedback loops \cite{sriraman_topology_2021, Cranford2013Materiomics_AM, chen_molecular_2024, aldeghi_graph_2022, ferraz_de_arruda_contagion_2024,Marom2025BioMaterialIntelligence,Cranford2013Materiomics_AM}. Representing such systems with a standard graph where edges denote pairwise interactions can fail to capture the full complexity of these multi-component and interdependent interactions. 

\begin{figure}
    \centering
    \includegraphics[width=1\linewidth]{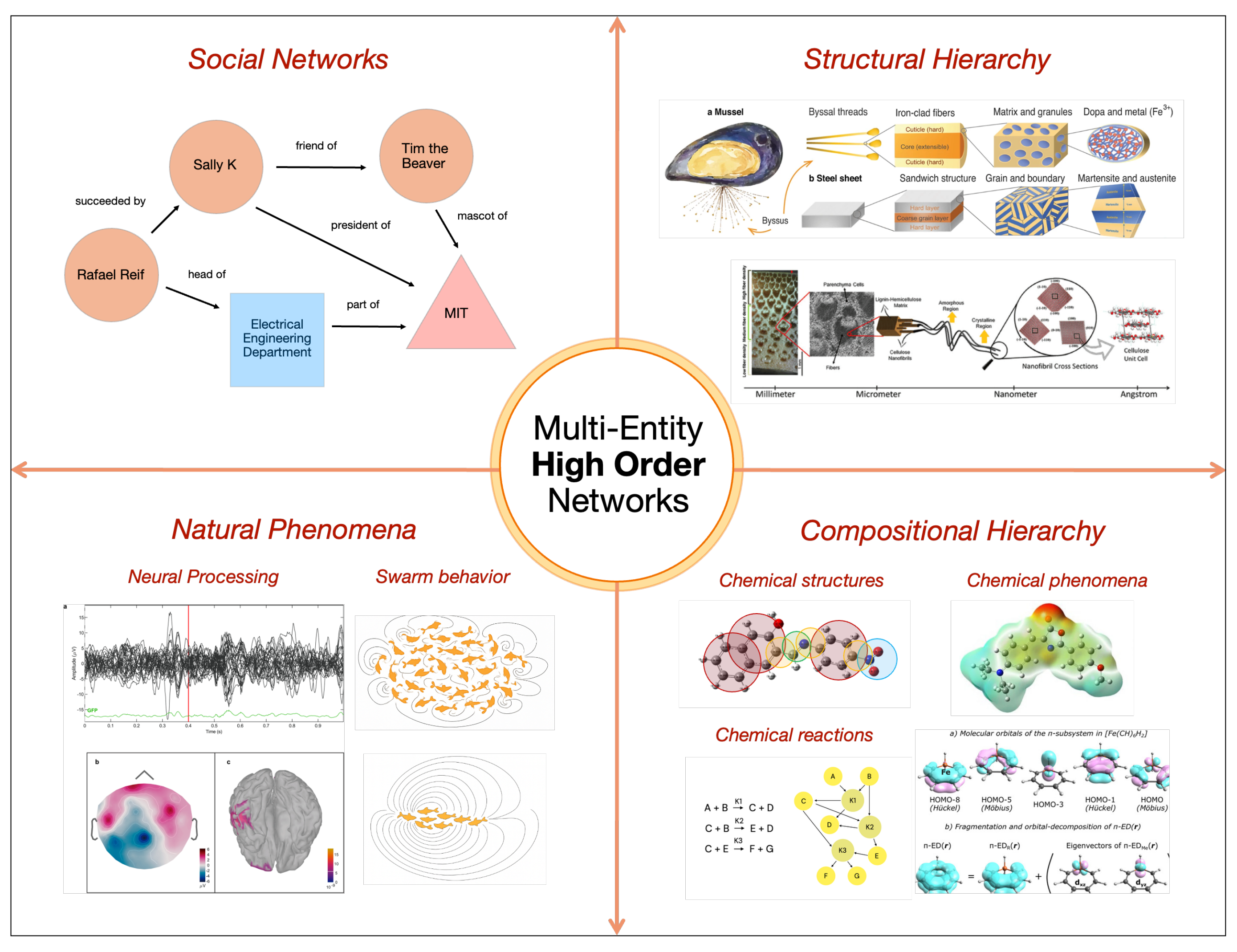}
    \caption{High-order multi-entity networks organized along two conceptual axes. The horizontal axis contrasts living-agent systems (left) with material systems (right), and the vertical axis separates macro-organizational structure (top) from micro-level mechanistic interactions (bottom). The four quadrants illustrate: (a) social networks, where roles and relationships create multiway associations; (b) biological emergence, such as neural processing and swarm behavior driven by local interaction rules. Reprinted with permission from Reference \cite{liu_open_2024} Licensed under CC-BY; (c) structural hierarchy in materials, where nested architectures across length scales cooperatively determine performance. As demonstrated in studies of mussel byssal threads and bamboo culms, structural hierarchy enables disparate materials to achieve exceptional strength and robustness. Reprinted with permission from Reference \cite{libonati_advanced_2017, cao_nature-inspired_2018} Licensed under CC-BY; and (d) compositional hierarchy, exemplified in chemistry where molecular bonding, electron delocalization, and reaction networks intrinsically involve simultaneous interactions among multiple atoms, electrons, and reactants rather than isolated pairwise relationships. Reprinted with permission from Reference \cite{nazli_tuning_2023, szczepanik_electron_2019} Licensed under CC-BY}
    \label{fig:highorder}
\end{figure}

In Figure \ref{fig:highorder}, we illustrate four major classes of multi-entity interactions organized along two orthogonal conceptual axes. The horizontal axis distinguishes living-agent systems (left), in which interacting units exhibit biological or cognitive agency, from material systems (right) where interactions are governed by physical or chemical constraints. The vertical axis separates macro-organizational hierarchies (top), where coordinated interactions produce large-scale collective organization, from micro-level mechanistic hierarchies (bottom), where emergent behavior arises from local interaction rules among many individual units. The top-left quadrant highlights social networks, where interpersonal roles and relationships generate multiway associations that extend beyond isolated pairwise ties \cite{bassett_network_2017, barabasi_network_2016, coulson_strength_2010, wasserman_social_1994}. The bottom-left quadrant depicts natural biological emergence, including neural computation \cite{sporns_networks_2011} and swarming behavior, where bees and fish coordinate through local signaling, motion cues, and hydrodynamic sensing to generate synchronized movement and collective decision-making \cite{seeley_wisdom_2009, camazine_self-organization_2003}. On the material side, the top-right quadrant represents structural hierarchy in biological and engineered materials, exemplified by mussel byssal threads \cite{ehrlich_marine_2019} and cellulosic plant tissue \cite{eichhorn_review_2010}. Here, fibers, grains, interfaces, and nested architectural features across macro-, micro-, and nanoscales interact cooperatively to establish global mechanical performance, including toughness, extensibility, and durability. Such properties cannot be understood by examining a single structural level in isolation. The bottom-right quadrant reflects compositional hierarchy in chemistry, where bonding interactions, reaction networks, and electron delocalization phenomena \cite{carey_advanced_2007} intrinsically require simultaneous participation of multiple atoms, electrons, or reactants \cite{atkins_molecular_2011}. Taken together, the four quadrants demonstrate that high-order, multi-entity interactions underlie living-agent systems, material structure, and chemical composition, and that these systems are not adequately represented when reduced to classical pairwise graphs.

\begin{figure}
    \centering
    \includegraphics[width=1\linewidth]{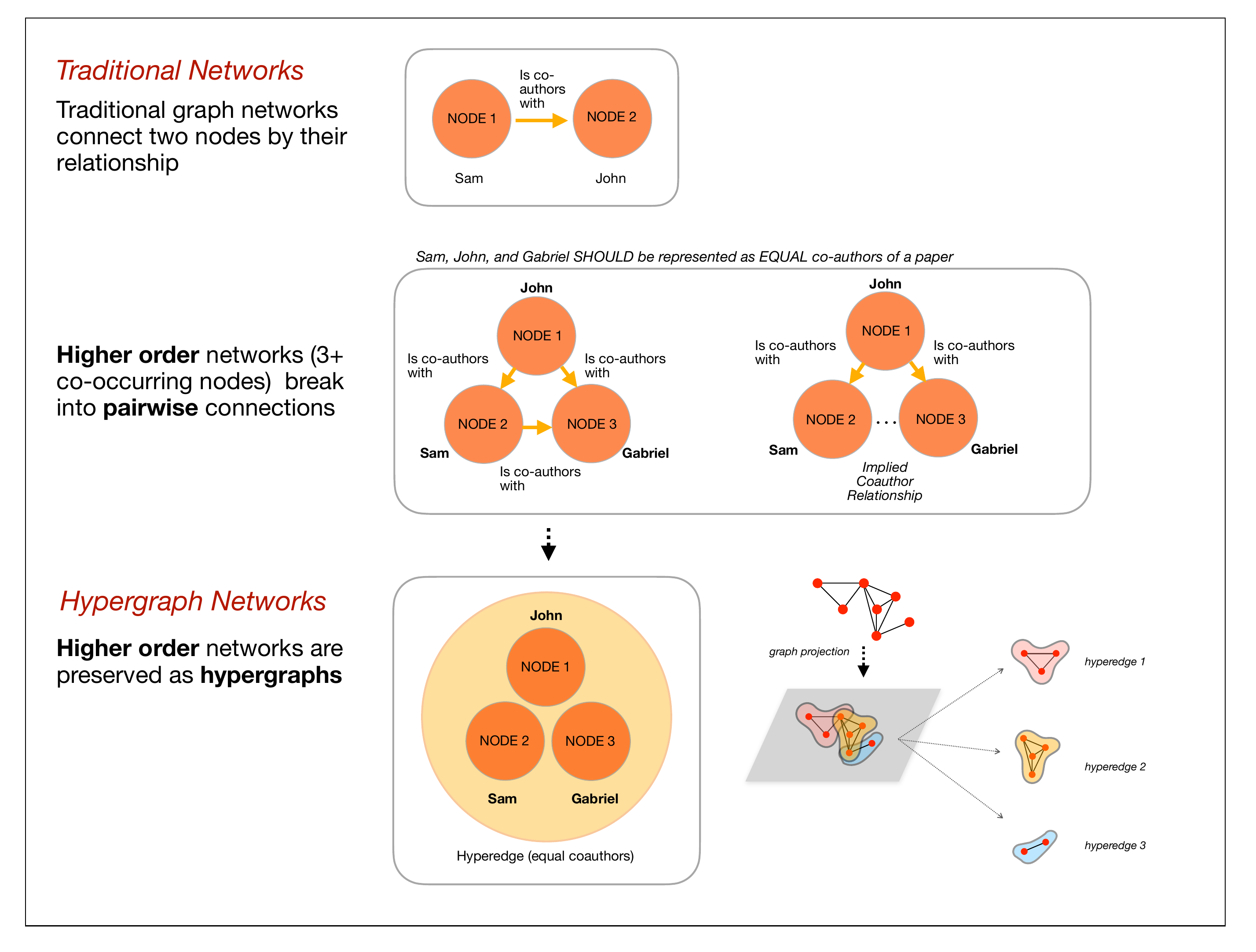}
    \caption{Comparison between traditional pairwise graph representations and hypergraph representations for higher-order relationships. While traditional graphs decompose multi-entity interactions into implied pairwise edges, hypergraphs preserve co-occurrence relationships as a single higher-order hyperedge, enabling more faithful representation of multiway associations. The example illustrates equal co-authorship, where decomposing a three-author contribution into pairwise connections distorts the equality of all coauthors and obscures the true relational structure underlying the paper.}
    \label{fig:hypergraph_idea}
\end{figure}

The key idea proposed in this paper is that we introduce hypergraphs as a more natural representation, since they capture relationships among multiple entities simultaneously, depicted in Figure \ref{fig:hypergraph_idea}. Just as a chemical reaction links several molecules in a single event, a hyperedge connects multiple vertices at once, allowing triadic, tetradic, and higher-order interactions to be modeled \cite{chang_hypergraph_2024, konstantinova_application_2001}. 

In this study, we analyze the reasoning capabilities of LLMs applied to hypergraphs and evaluate how effectively they retrieve and infer information from higher-order relational representations. We construct a hypergraph from a large corpus of scientific literature on biocomposite scaffolds and leverage hypergraph pattern mining to reveal how knowledge in the field is organized. By representing multi-entity statements as native higher-order interactions rather than collapsing them into pairwise edges, the hypergraph supports global corpus analysis, enabling the discovery of recurring higher-order motifs, densely interconnected communities, and emergent mechanistic patterns that are directly encoded in the literature \cite{juul_hypergraph_2024}. Hypergraph mining toolkits extend this capability to multi-scale analysis, including motif statistics, higher-order clustering, rich-club detection, and generative structure models applicable to large corpora \cite{lee_survey_2025}. In tandem, local community detection based on folded or projected subgraphs enables targeted exploration around seed concepts, iteratively expanding dense hyperedge neighborhoods and exposing topic-specific micro-fields, methodological subdomains, and mechanistic adjacencies otherwise obscured in pairwise graphs \cite{zhang_local_2025}. These tools expose higher-order structure and relationships in the corpus that pairwise graphs cannot recover.

We perform inference on the constructed hypergraph within a multi-agent framework, where only one agent has direct access to the hypergraph and the remaining agents specialize in domain-specific reasoning. The hypergraph-equipped agent identifies candidate paths or mechanistic chains, and the specialized agents sequentially elaborate, critique, and refine these into concrete hypotheses. Such a division of responsibilities is analogous to collaborative research groups, in which multiple experts provide complementary perspectives that collectively drive a broader scientific conclusion. 

By pooling diverse agent capabilities around a shared hypergraph substrate, the system can uncover relationships, alternative mechanisms, and design opportunities that are difficult for a single model to infer. As a demonstration, we apply a multi-agent framework to perform inference over our hypergraph constructed from biocomposite scaffold literature, where multiple constituent ingredients jointly define the resulting material. Agents leverage the higher-order structure to generate novel composite formulations and candidate experimental protocols. 

\subsection{Related Works}

Recent advances in $n$-ary information extraction and hypergraph reasoning have motivated richer, higher-order knowledge representations for complex scientific corpora. Text2NKG proposes fine-grained $n$-ary relation extraction to construct relational knowledge graphs directly from natural language without collapsing multi-entity interactions into binary triples \cite{luo_text2nkg_2024}, while MIRROR introduces a unified multi-slot tuple formulation for diverse information extraction tasks and decodes all spans using a non-autoregressive cyclic graph structure, improving versatility across complex IE and reading comprehension settings \cite{zhu_mirror_2023}. Structured knowledge representation systems such as HyperG further demonstrate that tabular or semi-structured data can be encoded as hypergraphs, improving QA accuracy and fact verification with LLMs by retaining n-ary, multi-row dependencies \cite{huang_hyperg_2025}. 

Hypergraph-based reasoning has been explored in multiple application areas. The Hypergraph Transformer models multi-hop dependencies for visual question answering \cite{heo_hypergraph_2022}. LLM4Hypergraph provides the first comprehensive benchmark for higher-order reasoning in LLMs and shows that hypergraph-aware prompting strategies, including Hyper-BAG (Build-a-Hypergraph Prompting) and Hyper-COT (Hypergraph Chain of Thought), improve structural inference accuracy across synthetic and real-world hypergraphs by helping LLMs visualize hypergraph architecture and perform stepwise connectivity analysis \cite{feng_beyond_2024}. Hyper-COT is a task-oriented prompting method tailored to benchmark hypergraph classification and structural reasoning, while Hyper-BAG supports mental construction of hyperedges and vertex relationships during inference. In contrast, hypergraph-inspired chain-of-thought (COT) approaches introduced by Yao et al. represent the reasoning trajectory itself as transitions among multi-entity sets, preserving higher-order relational context and improving deductive coherence \cite{yao_thinking_2023}. 

Hypergraph structures have also been used to enhance communication and coordination in multi-agent systems, with HyperComm modeling coalition communication via hyperedges to improve task performance and cooperative decision-making in reinforcement learning settings \cite{zhu_hypercomm_2024}. For retrieval-augmented generation, Hyper-RAG uses hypergraph structure over knowledge corpora to improve factual grounding and reduce hallucinations, outperforming GraphRAG and LightRAG on complex QA tasks \cite{feng_hyper-rag_2025}. However, existing approaches typically either extract n-ary relations, treat hypergraphs as auxiliary structures for QA or communication, or evaluate synthetic hypergraph reasoning in isolation; none construct large-scale, domain-native hypergraphs from real scientific corpora, systematically mine their higher-order organization, and evaluate multi-agent LLMs performing mechanistic inference and hypothesis generation directly within this representational substrate. Our work fills this gap by treating the hypergraph as the primary scientific knowledge structure rather than a downstream feature, enabling automated reasoning, materials discovery, and mechanistic hypothesis formation that are inherently higher-order and irreducible to pairwise graphs.

\subsection{Hypergraph Preliminary: Representing Higher-Order Relations in Graph Structure}

We consider hypergraph theory with this context and investigate the limitations of projecting higher order relations to dyadic projections used in standard graph theory. 

A hypergraph \( H = (V, E) \) consists of a set of nodes \( V \) and a set of hyperedges \( E \), where each hyperedge \( e \in E \) is a subset of \( V \), i.e., \( e \subseteq V \). Each node \( v \in V \) must appear in at least one hyperedge, meaning \( V = \bigcup_{e \in E} e \).

Unless stated otherwise:
\begin{itemize}
  \item Hyperedges are not weighted.
  \item Duplicate hyperedges are not considered.
  \item Each hyperedge contains at least two nodes, i.e., \( |e| \geq 2 \).
\end{itemize}

Unlike pairwise graphs, where each edge connects exactly two nodes, a hyperedge in a hypergraph can connect any number of nodes. The size of a hyperedge \( e \) is defined as the number of nodes it contains, denoted \( |e| \).

For a hypergraph \( H = (V, E) \), the \textbf{degree} of a node \( v \in V \), denoted \( d(v; H) \), is the number of hyperedges that include \( v \), i.e.,
\[
d(v; H) = \left| \{ e \in E : v \in e \} \right|.
\]

Given a subset \( S \subseteq V \), the \textbf{volume} of \( S \) in \( H \), written \( \text{vol}(S; H) \), is the sum of the degrees of all nodes in \( S \):
\[
\text{vol}(S; H) = \sum_{v \in S} d(v; H).
\]

\subsection*{Induced Subhypergraphs}

A hypergraph \( H' = (V', E') \) is called a subhypergraph of another hypergraph \( H = (V, E) \) if \( E' \subseteq E \).

The induced subhypergraph of \( H \) on a subset \( S \subseteq V \) is denoted by \( H[S] \), and defined as:
\[
H[S] = (S, \{ e \in E : e \subseteq S \}).
\]

\subsection*{Incidence Matrices}

A straightforward matrix representation of a hypergraph \( H = (V, E) \) is the incidence matrix \( M_I(H) \), which has \( |V| \) columns and \( |E| \) rows. Each entry indicates whether a node \( v \in V \) is part of a hyperedge \( e \in E \), defined as:
\[
M_I(v, e; H) =
\begin{cases}
1, & \text{if } v \in e, \\
0, & \text{otherwise}.
\end{cases}
\]

\subsection*{Paths and Connectivity}

A path is a sequence of hyperedges \( (e_1, e_2, \ldots, e_\ell) \) of length \( \ell \in \mathbb{N} \), such that \( e_i \cap e_{i+1} \neq \emptyset \) for all \( i \in [\ell - 1] \).

A hypergraph \( H = (V, E) \) is connected if every pair of nodes \( v_1, v_2 \in V \) can be joined by such a path, with \( v_1 \in e_1 \) and \( v_2 \in e_\ell \). Among all such paths, the shortest paths (possibly multiple of the same length) are the ones with the minimum number of hyperedges.

A subset of hyperedges \( E' \subseteq E \) is said to be connected if the hypergraph \( H' = (V' = \bigcup_{e \in E'} e, E') \) is connected.

\subsection*{Formal Dyadic Projections}

It is of note hypergraphs can be naturally approximated by a pairwise graph through dyadic projections.
 Two common ways of dyadic projections are clique expansions and star expansions. By default, we consider unweighted dyadic projections, and explicit clarification will be added for exceptions.

\begin{mdframed}[linewidth=1pt, linecolor=black!60, backgroundcolor=gray!5]
\begin{defn}[Clique Expansion]
Given a hypergraph $\mathcal{H} = (V, E)$, the clique expansion of $\mathcal{H}$ is the (pairwise) graph $G_{ce}(\mathcal{H}) = (V, E_{ce})$, where each hyperedge is replaced by a complete graph (clique) over its nodes.

\medskip

\noindent\textit{Formally:}
\[
E_{ce} = \bigcup_{e \in E} \binom{e}{2}.
\]

That is, for each hyperedge $e \in E$, we create pairwise edges between all pairs of nodes in $e$.
\end{defn}
\end{mdframed}

While clique expansions offer a convenient pairwise approximation of hypergraphs, they fail to preserve the higher-order interactions inherent to the original structure. Consequently, critical relational information is lost in the projection. This limitation has motivated ongoing research in the area of hypergraph reconstruction, which seeks to infer or recover the underlying higher-order connectivity from observed pairwise data \cite{young_hypergraph_2021, lizotte_hypergraph_2023, wang_graphs_2024}. In general, the original hypergraph cannot be uniquely reconstructed from its clique expansion, as the projection is not invertible and multiple higher-order configurations may correspond to the same set of pairwise relationships.

\paragraph{Example (Clique Expansion):} Assume the following that Sally, Bob, David, and Ella are equal co-authors of a paper. The clique expansion represents this group as all possible pairwise connections:

\begin{verbatim}
{"Sally"} {"is co-authors with"} {"Bob"}
{"Sally"} {"is co-authors with"} {"David"}
{"Sally"} {"is co-authors with"} {"Ella"}
{"Bob"}   {"is co-authors with"} {"David"}
{"Bob"}   {"is co-authors with"} {"Ella"}
{"David"} {"is co-authors with"} {"Ella"}
\end{verbatim}

One advantage of the pairwise graph representation is that it makes higher-order relationships appear fully connected where each entity is linked to all others, resulting in a network where every node has the same degree and no entity is more ``central'' than another. However, this approach comes with significant limitations. First, it leads to a combinatorial explosion of edges: representing a single group interaction among \( n \) entities requires 
\( (n(n-1))/2 \) pairwise edges. Second, important contextual information is lost. In this example, the identity of the paper on which the authors collaborated is not preserved, making it impossible to distinguish between repeated collaborations and unique group efforts. Furthermore, the cohesive nature of group collaboration is obscured, as the pairwise representation implies several independent interactions rather than a unified co-authorship. Third, it artificially inflates structural metrics: the clustering coefficient, which measures the tendency for nodes' neighbors to be interconnected (indicating tightly knit groups), becomes inflated. Similarly, degree assortativity, which assesses whether nodes tend to connect to others with similar numbers of connections, is also exaggerated due to the imposed full connectivity.

\begin{mdframed}[linewidth=1pt, linecolor=black!60, backgroundcolor=gray!5]
\begin{defn}[Star Expansion]
Given a hypergraph $\mathcal{H} = (V, E)$, the star expansion of $\mathcal{H}$ is the bipartite graph $G_{se}(\mathcal{H}) = (V \cup E, E_{se})$, where hyperedges become nodes and edges connect each original node to the hyperedges that contain it.

\medskip

\noindent\textit{Formally:}
\[
E_{se} = \{ (v, e) \in V \times E : v \in e \}.
\]
\end{defn}
\end{mdframed}

\paragraph{Example (Star Expansion):} Assume the following that Sally, Bob, David, and Ella are equal co-authors of a paper. The star expansion introduces a new node representing the paper itself (e.g., \texttt{Paper1}) and connects each author to this node:

\begin{verbatim}
{"Sally"} {"is a coauthor to"} {"Paper1"}
{"Bob"}   {"is a coauthor to"} {"Paper1"}
{"David"} {"is a coauthor to"} {"Paper1"}
{"Ella"}  {"is a coauthor to"} {"Paper1"}
\end{verbatim}

The star expansion preserves the full incidence information of the original hypergraph by explicitly introducing hyperedges as nodes (e.g., papers). This maintains the identity of each group interaction, enabling us to distinguish which authors collaborated on which paper and how frequently. It also avoids the combinatorial explosion of edges found in clique expansions, requiring only \( n \) edges for a group of \( n \) co-authors instead of \( \frac{n(n-1)}{2} \), and does not artificially inflate network metrics such as clustering coefficient or degree assortativity.

However, the representation introduces a new node type which in this case is the paper, shifting the structure from an author-author network to an author-paper network. This abstraction can obscure direct collaboration patterns between authors, making certain queries like “which authors frequently collaborate together?” more difficult to answer without projecting back to a full clique structure. Moreover, standard graph algorithms do not distinguish between node types, which may yield misleading results unless careful modeling or annotation is applied. While the group context is technically preserved, it is not directly encoded in the structure of the graph without further projection or interpretation.

\subsection*{Informal Dyadic Projections}

Hypergraphs can additionally be projected into non-standard but interpretable forms of network representation, often used heuristically or for simplifying higher-order relationships. This is typically observed of LLMs tasked with extracting knowledge triples due to the efficiency of the token usage unless explicitly instructed to do otherwise. We provide formal-style definitions for three informal dyadic projections: the collapsed, cyclic implicit, and chain implicit graph representations.

\vspace{4mm}

\begin{mdframed}[linewidth=1pt, linecolor=black!60, backgroundcolor=gray!5]
\begin{defn}[Collapsed Representation]
Given a hyperedge $e = \{v_1, v_2, \ldots, v_n\}$, the collapsed representation maps the entire set $e$ to a single edge, treating the group as an atomic unit. This edge is labeled or typed to carry information about the relationship.

\medskip

\noindent\textit{Formally:} This results in a graph $G_{col} = (E_{group}, L)$, where each node in $E_{group}$ is a subset of original nodes $\subseteq V$, and $L$ contains metadata.
\end{defn}
\end{mdframed}

\vspace{4mm}

\noindent\textbf{Example (Collapsed Representation):} 
Assume that Sally, Bob, David, and Ella are equal co-authors of a paper. The collapsed representation models this entire group interaction as a single hyperedge:

\begin{quote}
\texttt{\{"Sally, Bob, David, and Ella"\} \{"are equal co-authors of"\} \{"Paper 1"\}}
\end{quote}

This representation preserves the identity of the collaborative event such as the co-authorship of a paper, while maintaining minimal structural complexity, as a single edge is sufficient to encode the entire group interaction. However, this abstraction comes at the cost of granularity. Individual contributors are not represented as distinct nodes, thereby precluding the ability to track their independent roles or participation across multiple interactions. For example, if a given author engages in additional collaborations outside the group, such relationships cannot be separately captured or analyzed within this model. This representation is most appropriate when the analytical focus is on the collective act or artifact itself, rather than on the individual entities involved or their broader interaction patterns.

\begin{mdframed}[linewidth=1pt, linecolor=black!60, backgroundcolor=gray!5]
\begin{defn}[Cyclic Implicit Representation]
Given a hyperedge $e = \{v_1, v_2, \ldots, v_n\}$, the cyclic implicit representation models the group interaction as a closed loop (cycle) among the entities. Each node is connected to two neighbors in the group, forming the cycle graph $C_n$. 

\medskip

\noindent\textit{Formally:} This yields a graph $G_{cyc} = (V, E_{cyc})$, where
\[
E_{cyc} = \{(v_i, v_{i+1}) \mid 1 \leq i < n\} \cup \{(v_n, v_1)\}.
\]
\end{defn}
\end{mdframed}

\paragraph{Example (Cyclic Implicit Representation):} Assume the following that Sally, Bob, David, and Ella are equal co-authors of a paper. The cyclic implicit representation connects each person to two others in a ring-like structure:

\begin{verbatim}
{"Sally"} {"is co-authors with"} {"Bob"}
{"Bob"}   {"is co-authors with"} {"David"}
{"David"} {"is co-authors with"} {"Ella"}
{"Ella"}  {"is co-authors with"} {"Sally"}
\end{verbatim}

This representation reduces edge complexity relative to the full clique expansion while preserving individual node identities. Each author node maintains a uniform degree centrality of 2, where degree centrality is the number of edges incident to a node, and the resulting topology forms a closed loop that implicitly suggests group participation.

Nonetheless, the structure encodes only local pairwise interactions and fails to capture the full higher-order relationship. The identity of the collaborative artifact, such as the co-authored paper, is not retained, and the cohesive nature of the group interaction is not explicitly represented in the graph topology. As a result, the presence of a joint collaboration must be inferred rather than directly observed. This sparsified topology can also distort centrality-based measures. Betweenness centrality, defined as the fraction of shortest paths in the network that pass through a given node, may become disproportionately elevated for authors positioned on bridging paths between otherwise disconnected nodes, even when their actual participation in the underlying group is no different from others. Such distortions can lead to misleading conclusions about an individual’s importance or influence within the collaboration network.

\begin{mdframed}[linewidth=1pt, linecolor=black!60, backgroundcolor=gray!5]
\begin{defn}[Chain Implicit Representation]
Given a hyperedge $e = \{v_1, v_2, \ldots, v_n\}$, the chain implicit representation models the group interaction as a simple path graph $P_n$ over the entities, connecting each node to its immediate neighbor in a linear sequence. 

\medskip

\noindent\textit{Formally:} This results in a graph $G_{chain} = (V, E_{chain})$, where
\[
E_{chain} = \{(v_i, v_{i+1}) \mid 1 \leq i < n\}.
\]
\end{defn}
\end{mdframed}

\paragraph{Example (Chain Implicit Representation):} Assume the following that Sally, Bob, David, and Ella are equal co-authors of a paper. The chain implicit representation links them in a sequential path:

\begin{verbatim}
{"Sally"} {"is co-authors with"} {"Bob"}
{"Bob"}   {"is co-authors with"} {"David"}
{"David"} {"is co-authors with"} {"Ella"}
\end{verbatim}

This model reduces the number of pairwise edges compared to the full clique expansion, using only \( n - 1 \) edges for \( n \) authors. It also retains individual node identities, preserving author-level granularity. However, the full higher-order relationship is not explicitly encoded. The identity of the collaborative artifact (e.g., the co-authored paper) is lost, and there is no structural indication that the authors participated as a cohesive group. Moreover, the topology introduces distortions in network centrality metrics. Sally and Ella each have only one edge, resulting in a lower degree centrality (1), while Bob and David have two edges and thus appear more central, despite equal participation in the collaboration. Betweenness centrality is also skewed: Bob and David lie on the shortest paths between otherwise disconnected pairs, such as Sally and Ella, whose shortest path spans three edges. This incorrectly inflates the perceived importance of intermediaries and may misrepresent the closeness of the group’s collaboration.

\subsection*{Nested Hypergraphs}

A nested hypergraph generalizes the classical notion of a hypergraph by permitting \emph{hyperedges to contain other hyperedges}. This structure introduces a hierarchy among hyperedges, where an \emph{outer} hyperedge may contain one or more \emph{inner} hyperedges, potentially over multiple levels of nesting.

\textbf{Informal Definition.} Although no universally accepted formal definition of nested hypergraphs exists, an implicit and commonly used characterization is as follows:
\[
\text{A hyperedge } e_1 \in E \text{ is \emph{nested} in hyperedge } e_2 \in E \text{ if } e_1 \subseteq e_2.
\]

This condition allows hyperedges to act as both sets of nodes and containers of other hyperedges, forming hierarchical relationships.

Nested hypergraphs are well-suited for modeling systems with inherently hierarchical or multiscale interactions. Applications include, but are not limited to:
\begin{itemize}
    \item Complex pathway modeling 
    \item Network analysis with group-subgroup dynamics
    \item Hierarchical entity representation in natural language processing
\end{itemize}

\section{Results and Discussion}\label{sec3}

With these concepts in place we know proceed to apply this framework to a real-world example. 

\subsection{Construction of Hypergraph}

We first present an incremental procedure for constructing scientific knowledge hypergraphs detailed in Algorithm~\ref{alg:llm_hypergraph} (additional details, see Materials and Methods).

\subsubsection{Document Preprocessing and Extraction}
Each document $d_i \in \mathcal{D}$ is partitioned to improve contextual resolution while reducing token usage into sections of 10,000 characters with zero overlap using recursive text splitting. Each chunk receives a unique identifier for provenance tracking. Optional preprocessing includes content distillation such as LLM-generated summaries preserving facts while removing citations, and figure analysis where vision-language models are applied for extracting structured information from images and tables. 

We employ a two-pass strategy to collect multi-entities in text in order to balance precision and recall. 

\begin{mdframed}[linewidth=1pt, linecolor=black!60, backgroundcolor=gray!5]
\textbf{Dual-Pass Strategy for Multi-Entity Extraction}

\medskip

\noindent\textbf{Pass 1 -- Exact grammatical extraction:} Identifies explicit Subject--Verb--Object triples with verbatim predicates including prepositions (``used for,'' ``employed in,'' ``leads to''), handling transitive verbs, copular predicates, prepositional predicates, and relative clauses. A pre-pass detects composite materials (``chitosan/hydroxyapatite nanocomposite'') via delimiters (``/'', ``-'', ``and''), emitting special ``compose'' relations.

\medskip

\noindent\textbf{Pass 2 -- Conservative semantic completion:} Recovers implicit relationships within sentence boundaries via transformations: nominalizations to light verbs (``fabrication of X'' $\to$ relation: ``fabricate''), appositions to identity relations (``Collagen, a structural protein'' $\to$ relation: ``is''), purpose phrases (``X for Y'' $\to$ relation: ``used for''), and causal connectives (``thereby'' $\to$ ``leads to''). Node labels remain verbatim; only relations may be abstracted.
\end{mdframed}

To ensure precise entity extraction, demonstratives (``this material'') are resolved to specific antecedents within 1--2 sentences, while generic unmodified terms (``material,'' ``polymer,'' ``method,'' ``device'') are omitted unless accompanied by modifiers (e.g., ``nHAp-based polymer nanocomposite scaffold''). The LLM returns JSON: \texttt{\{source: [entities], relation: string, target: [entities]\}}, naturally handling n-ary relationships.

\subsubsection{Hypergraph Construction}
From extracted events $\mathcal{R}_i$, we build document-level hypergraph $\mathcal{H}_i = (V_i, E_i)$ where each event becomes a hyperedge: $E_i = \{\text{source} \cup \text{target} : (\text{source}, \text{target}, r, c) \in \mathcal{R}_i\}$. Unlike pairwise graphs, hyperedges naturally capture n-ary relationships without auxiliary nodes. Each hyperedge is labeled with relation $r$ and chunk identifier $c$. A synchronized DataFrame $\mathcal{T}_i$ maps edges to source lists, target lists, relations, and originating chunks.

After processing each document, we immediately merge $\mathcal{H}_i$ into global hypergraph $\mathcal{H}$ via HyperNetX union. This provides memory efficiency by discarding intermediate subgraphs and early cross-document integration improving subsequent deduplication via enriched degree information. Provenance accumulates: $\mathcal{T} \leftarrow \mathcal{T} \cup \mathcal{T}_i$.

\subsubsection{Hypergraph Cleaning}
Hypergraph cleaning is required often when scientific terminology varies across papers (``PLA,'' ``polylactic acid,'' ``poly(lactic acid)''). Every $f = 10$ documents, we perform embedding-based deduplication. For new nodes $v_j \in V_i$, we compute embeddings $\mathbf{v}_j \leftarrow \phi(v_j)$ using a \emph{nomic} sentence embedding model and find the pairwise cosine similarities $S_{jk}$ within this vector space. Nodes with $S_{jk} \geq 0.95$ form similarity graph $\mathcal{G}_{\text{sim}}$, whose connected components yield equivalence classes $\mathcal{C}$.

For each equivalence class $C \in \mathcal{C}$, we select a representative node using a degree-based heuristic: $\rho(C) = \arg\max_{v \in C} \text{deg}(v, E)$, where $\text{deg}(v, E)$ denotes the number of hyperedges incident to node $v$. This strategy retains the most frequently-referenced terminology under the assumption that high-degree nodes represent central, well-established concepts in the literature. For instance, if the cluster contains \{``PLA'', ``polylactic acid'', ``poly(lactic acid)''\} with degrees 47, 23, and 8 respectively, we select ``PLA'' as the canonical label.

The merging process executes four synchronized operations to maintain structural and semantic consistency:

\begin{mdframed}[linewidth=1pt, linecolor=black!60, backgroundcolor=gray!5]
\textbf{Synchronized Node Merging Operations}

\medskip

\noindent\textbf{(1) Text aggregation:} We preserve complete provenance by merging all associated text chunks from merged nodes: $\text{text}(\rho(C)) \leftarrow \bigcup_{v \in C} \text{text}(v)$. This ensures no source material is lost such that the representative node inherits all contextual information from its synonyms.

\medskip

\noindent\textbf{(2) Dataframe synchronization:} We apply the node mapping $\sigma: V \to V$ defined by $\sigma(v) = \rho([v])$ (where $[v]$ denotes the equivalence class containing $v$) to all source and target columns in the provenance dataframe $\mathcal{T}$. For example, all rows containing ``polylactic acid'' in either source or target columns are updated to ``PLA''.

\medskip

\noindent\textbf{(3) Hypergraph reconstruction:} We reconstruct the incidence dictionary by applying $\sigma$ to all node labels within hyperedges. During this process, some hyperedges may become self-loops meaning edges where all nodes collapse to a single representative. For instance, if a hyperedge originally connected \{``polylactic acid'', ``PLA''\} and both map to ``PLA'', the resulting edge \{``PLA''\} is a degenerate single-node hyperedge. We remove such self-loops as they encode reflexive relationships (``PLA'' relates to ``PLA'') that provide no informational value. Similarly, any hyperedge in $\mathcal{T}$ where $s = t$ after mapping (source equals target) is filtered out.

\medskip

\noindent\textbf{(4) Embedding recomputation:} We update the embedding dictionary $\Phi$ by recomputing embeddings for all representative nodes using their canonical labels. This ensures subsequent similarity computations use the merged terminology rather than stale embeddings from individual synonyms. Embeddings for nodes in $\bigcup_{C \in \mathcal{C}} (C \setminus \{\rho(C)\})$ (i.e., all merged-away synonyms) are discarded.
\end{mdframed}

% \begin{algorithm}[t]
\begin{algorithm}
\caption{LLM-guided hypergraph construction with incremental merging}
\label{alg:llm_hypergraph}
\textbf{Input:} Document corpus $\mathcal{D} = \{d_1, \ldots, d_n\}$, LLM model $\mathcal{M} = (\mathcal{E}, \mathcal{R})$, embedding model $\phi$, similarity threshold $\theta = 0.95$, merge frequency $f = 10$\\
\textbf{Output:} Hypergraph $\mathcal{H} = (V, E)$, node embeddings $\Phi$, relationship dataframes $\mathcal{T}$\\
\textbf{Notation:} $\mathcal{H} = (V, E)$ hypergraph with nodes $V$ and hyperedges $E$; $\mathcal{H}_i = (V_i, E_i)$ document-level subgraph; $\mathcal{R}_i$ relationships as $(s,t,r,c)$ tuples (source, target, relation, chunk\_id); $S_{jk}$ cosine similarity between embeddings; $\mathcal{G}_{\text{sim}}$ similarity graph; $\mathcal{C}$ equivalence classes; $\rho(C)$ representative node (highest degree); $\sigma(v)$ node mapping function
\vspace{2mm}
\hrule
\vspace{2mm}
\begin{algorithmic}[1]
\STATE \emph{Phase 1: Incremental document-level hypergraph construction}
\vspace{0.5mm}
\STATE $\mathcal{H} \leftarrow (\emptyset, \emptyset)$, $\Phi \leftarrow \emptyset$, $\mathcal{T} \leftarrow \emptyset$
\STATE \textbf{for all} document $d_i \in \mathcal{D}$ \textbf{do}
\STATE \quad \textit{// Generate document-level subgraph}
\STATE \quad Chunk $d_i$ into pieces with size $c = 10{,}000$ chars
\STATE \quad \textbf{for each} chunk $k$ \textbf{do}
\STATE \quad \quad $\mathcal{R}_k \leftarrow \mathcal{R}(k)$ \hfill $\triangleright$ Extract (source[], target[], relation, chunk\_id) events
\STATE \quad \textbf{end for}
\STATE \quad $\mathcal{R}_i \leftarrow \bigcup_k \mathcal{R}_k$ \hfill $\triangleright$ Union all chunk events
\STATE \quad $\mathcal{T}_i \leftarrow \{(s, t, r, c) : s \in \text{source}, t \in \text{target}, (source, target, r, c) \in \mathcal{R}_i\}$
\STATE \quad $E_i \leftarrow \{\text{source} \cup \text{target} : (\text{source}, \text{target}, r, c) \in \mathcal{R}_i\}$ \hfill $\triangleright$ Create hyperedges
\STATE \quad $V_i \leftarrow \bigcup_{e \in E_i} e$ \hfill $\triangleright$ Collect all nodes
\STATE \quad $\mathcal{H}_i \leftarrow (V_i, E_i)$ with edge\_attr $\{e_k \mapsto \text{chunk\_id}(e_k)\}$
\vspace{1mm}
\STATE \quad \textit{// Incremental merge into global hypergraph}
\STATE \quad $V \leftarrow V \cup V_i$
\STATE \quad $E \leftarrow E \cup E_i$ 
\STATE \quad $\mathcal{T} \leftarrow \mathcal{T} \cup \mathcal{T}_i$ \hfill $\triangleright$ Accumulate chunk-level relationships
\vspace{1mm}
\STATE \quad \textit{// Conditional semantic merging}
\STATE \quad \textbf{if} $i \bmod f = 0$ \textbf{then}
\STATE \quad \quad Compute embeddings: $\mathbf{v}_j \leftarrow \phi(v_j)$ for new $v_j \in V_i$ 
\STATE \quad \quad Update $\Phi \leftarrow \Phi \cup \{\mathbf{v}_j\}$
\STATE \quad \quad Compute similarity: $S_{jk} \leftarrow \displaystyle\frac{\mathbf{v}_j^\top \mathbf{v}_k}{\|\mathbf{v}_j\| \|\mathbf{v}_k\|}$ for all pairs
\STATE \quad \quad $\mathcal{G}_{\text{sim}} \leftarrow \{(v_j, v_k) : S_{jk} \geq \theta\}$ \hfill $\triangleright$ Similarity graph
\STATE \quad \quad $\mathcal{C} \leftarrow \text{ConnectedComponents}(\mathcal{G}_{\text{sim}})$ \hfill $\triangleright$ Find equivalence classes
\STATE \quad \quad \textbf{for each} equivalence class $C \in \mathcal{C}$ \textbf{do}
\STATE \quad \quad \quad $\rho(C) \leftarrow \arg\max_{v \in C} \text{deg}(v, E)$ \hfill $\triangleright$ Keep higher-degree representative
\STATE \quad \quad \quad Merge texts: $\text{text}(\rho(C)) \leftarrow \bigcup_{v \in C} \text{text}(v)$ 
\STATE \quad \quad \quad Update $\mathcal{T}$: replace all $v \in C$ with $\rho(C)$ in source/target columns
\STATE \quad \quad \textbf{end for}
\STATE \quad \quad Apply node mapping: $\sigma(v) \leftarrow \rho([v])$ for all $v \in V$
\STATE \quad \quad $V \leftarrow \{\rho(C) : C \in \mathcal{C}\}$
\STATE \quad \quad $E \leftarrow \{\{\sigma(v) : v \in e, \sigma(v) \neq \text{null}\} : e \in E, |\{\sigma(v) : v \in e\}| > 1\}$
\STATE \quad \quad Remove self-loops: $\mathcal{T} \leftarrow \{(s,t,r,c) \in \mathcal{T} : s \neq t\}$
\STATE \quad \quad Update embeddings: $\Phi \leftarrow \{\phi(\rho(C)) : C \in \mathcal{C}\}$
\STATE \quad \textbf{end if}
\STATE \textbf{end for}
\STATE \textbf{return} $\mathcal{H} = (V, E)$, $\Phi$, $\mathcal{T}$
\end{algorithmic}
\end{algorithm}

\subsection{Analysis of Hypergraph of the Biocomposite Scaffold Corpora}

\begin{figure}[htbp]
    \centering
    \includegraphics[width=0.95\linewidth]{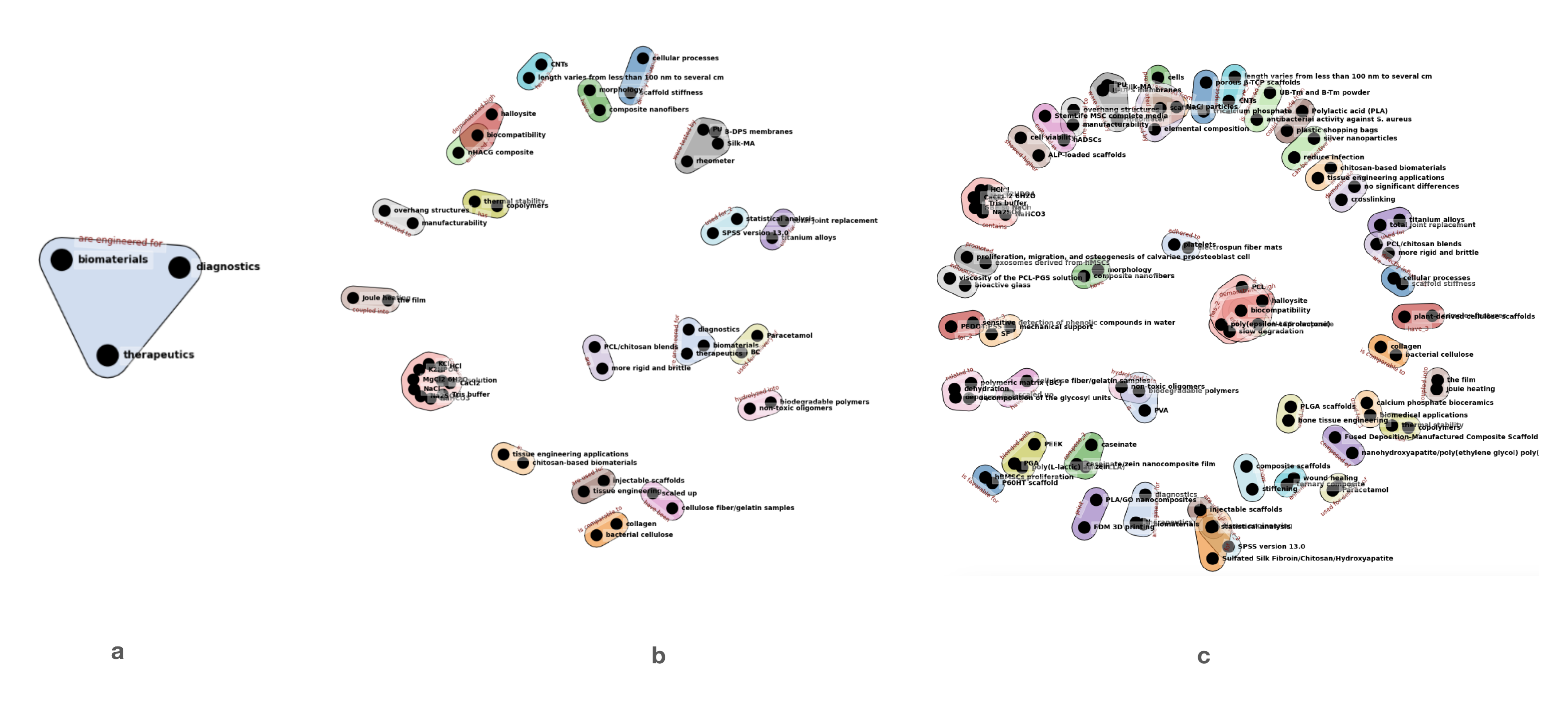}
    \caption{Randomly sampled hypergraph substructures from the gloabl \textit{Biocompatible Scaffold} hypergraph illustrating the evolution of higher-order topology as the number of hyperedges increases.
(a) A single hyperedge showing the higher-order relationship \textit{engineered for}, which jointly connects the nodes \textit{biomaterials}, \textit{diagnostics}, and \textit{therapeutics}. Although metadata is omitted for clarity, this hyperedge specifies node directionality, where \textit{biomaterials} points to both \textit{diagnostics} and \textit{therapeutics} through the edge \textit{engineered for}. (b) A random sample of 20 hyperedges from the global hypergraph. We observe one clustering event where the pairwise hyperedge \textit{nHACG composite} - \textit{has enhanced } - \textit{biocompatibility} overlaps with \textit{halloysite} - \textit{demonstrated high} - \textit{biocompatibility} from the node \textit{biocompatibility}. (c) A random sample of 50 hyperedges, showing more complex regional organization and richer connectivity patterns. As additional edges and nodes are introduced, hyperedge arrangements become increasingly structured, highlighting emergent topological behavior and concept clustering within the corpus.}
    \label{fig:labeled_edges}
\end{figure}

To study the structural behavior of the hypergraph more efficiently, we can visualize and examine random subgraphs sampled from the global network. Figure~\ref{fig:labeled_edges} displays random samples containing 1, 20, and 50 hyperedges with nodes and edges labeled as an exemplar to demonstrate the stored contents of the hypergraph. The singular hyperedge \textit{are engineered for} encapsulates nodes \textit{biomaterials}, \textit{diagnostics}, and \textit{therapeutics.} Metadata not shown in the figure contains the direction of the nodes, where \textit{biomaterials} points to \textit{diagnostics} and \textit{therapeutics} through by the edge \textit{engineered for.} As we expand the edge and node count, we can visualize how the hyperedge topology evolves.

\begin{figure}[htbp]
    \centering
    \includegraphics[width=0.9\linewidth]{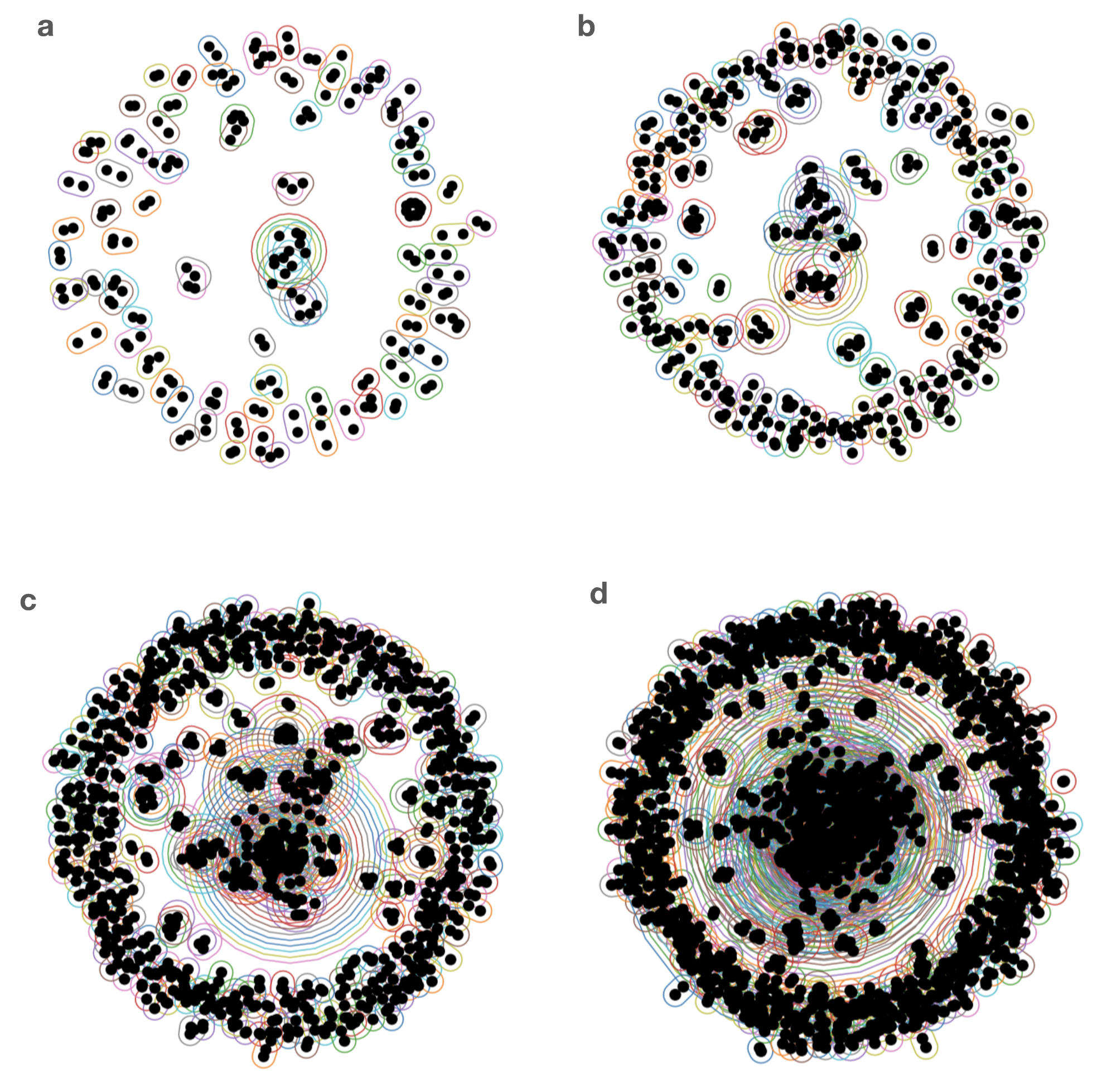}
    \caption{Subgraphs of the \textit{Biocomposite Scaffold } hypergraph generated with 100, 200, 500, and 1000 hyperedges. Increasing hyperedge count produces a clear core–periphery structure, with peripheral clusters integrating into a dense, highly shared central node set. Radial overlap among hyperedges at higher edge counts reflects strong multiway co-occurrence patterns in the underlying scientific corpus.}
    \label{fig:first_x_hypergraphs_2}
\end{figure}

\begin{table}[htbp]
\centering
\begin{tabular}{lp{3cm}}
\hline
\textbf{Property} & \textbf{Value} \\
\hline
\# Nodes & 161{,}172 \\
\# Edges & 320{,}201 \\
Avg edge size & 2.35 \\
Max edge size & 32 \\
Avg node degree & 4.68 \\
Max node degree & 11{,}157 \\
% \# Multi-edges (exact dup sets) & 58{,}997 \\
Max edge intersection size & 15 \\
\# pairs overlap $\ge 1$ & 22{,}106{,}585 \\
\# pairs overlap $\ge 2$ & 2{,}787{,}076 \\
\# pairs overlap $\ge 3$ & 212{,}355 \\
\hline
\end{tabular}
\caption{Summary of graph statistics}
\label{tab:fullgraph_stats}
\end{table}

To illustrate larger scales, we show Figure~\ref{fig:first_x_hypergraphs_2} of such random samples containing 100, 200, 500, and 1000 hyperedges. We omit node and edge labels for visual simplicity. As the number of sampled hyperedges increases, the topology transitions from a fragmented, loosely connected arrangement to a dense, cohesive core–periphery topology. At low edge counts, the system contains many small peripheral clusters and limited node sharing, reflecting sparse and localized co-occurrence patterns. With 200–500 hyperedges, these peripheral components progressively collapse inward as repeated hyperedges introduce overlapping node groups, revealing semantically meaningful relationships that begin to unify the graph. By 1000 hyperedges, a saturated central cluster emerges, characterized by heavy radial overlap among hyperedges and extensive reuse of a common set of high-frequency nodes. This progression indicates that the underlying scientific corpus contains strong higher-order relational structure: key concepts consistently appear together across multiple contexts, while lower-frequency terms gradually attach to this core as the sampling increases. The resulting topology highlights the limitations of pairwise graphs and demonstrates the value of hypergraphs for capturing multi-entity interactions fundamental to scientific reasoning.

Analyzing the graph holistically reveals several notable structural characteristics, summarized in Table \ref{tab:fullgraph_stats}. The hypergraph is large and sparse, containing 161{,}172 nodes and 320{,}201 hyperedges, yet its connectivity remains highly uneven. While the average node degree is only 4.68, the maximum degree reaches 11{,}157, indicating the presence of extreme hubs that dominate the connectivity landscape. A similar skew appears in edge composition: although the average edge size is just 2.35 nodes, the largest hyperedge contains 32 nodes, reflecting a small number of highly information-dense edges. The structure also exhibits substantial redundancy and overlap, with 58{,}997 exact duplicate hyperedges and a maximum edge–edge intersection size of 15. Pairwise co-occurrences extracted from all hyperedges highlight the combinatorial explosion characteristic of hypergraph decomposition: edges of size $\ge 1$ produce 22.1 million co-occurring pairs, but imposing stricter thresholds reduces this to 2.79 million pairs for $\ge 2$ co-occurrences and 212{,}355 pairs for $\ge 3$. This sharp decline underscores the noise present at low co-occurrence levels and the stronger structural coherence that emerges when filtering for more frequent or meaningful relationships.

\begin{figure}[htbp]
    \centering
    \includegraphics[width=1\linewidth]{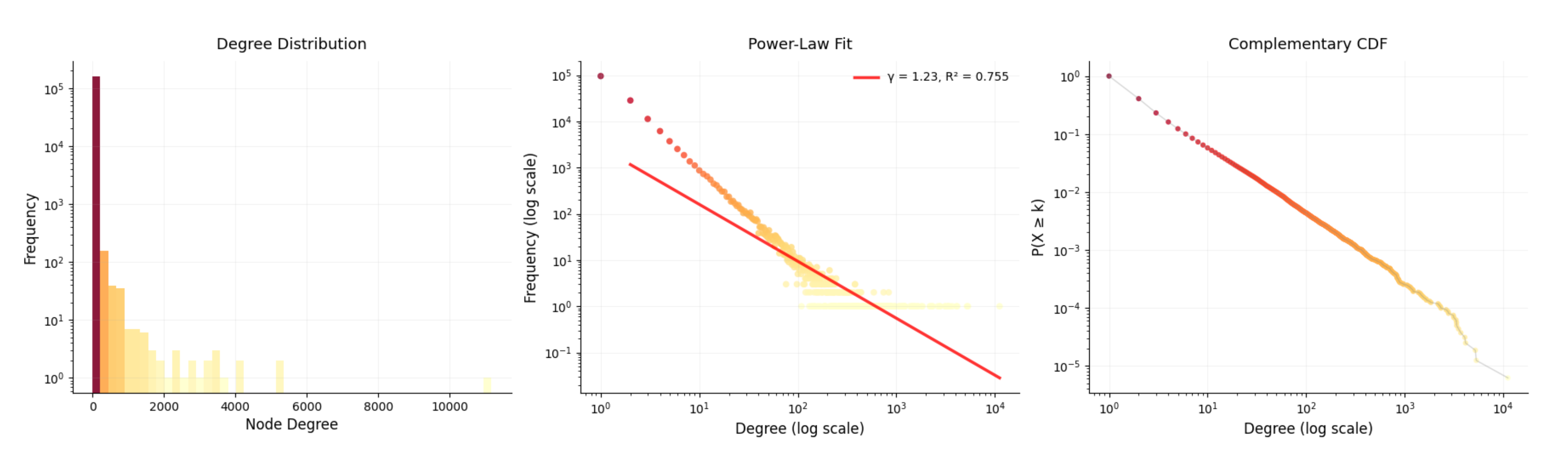}
    \caption{Node degree statistics and power-law behavior in the scientific hypergraph. The left panel shows the empirical degree distribution, where most nodes have low degree and a small subset form highly connected hubs. The middle panel depicts a log–log degree–frequency plot with a fitted power-law trend (y = 1.23, R² = 0.755), indicating heavy-tailed, scale-free connectivity. The right panel shows the complementary cumulative distribution function (CCDF), where the linear tail in log–log space further supports power-law scaling and reveals a small number of dominant semantic hubs within the corpus.}
    \label{fig:powerlaw}
\end{figure}

To better understand this core-periphery structure, we turn to the node-level patterns. The node degree distribution provides additional insight into the structure of the collected corpus. As shown in Figure~\ref{fig:powerlaw}, the distribution is heavy tailed and spans more than three orders of magnitude, with node degrees ranging from 1 up to over 11{,}000. The log–log frequency plot exhibits an approximate power law trend, with a fitted exponent near 1.23 and an associated coefficient of determination of about 0.755. Both the frequency plot and the complementary CCDF indicate that only a very small fraction of nodes occupy the high-degree regime, while the vast majority of nodes fall within the low-degree range below 20. The CCDF confirms this scaling behavior, although the rightmost portion of the tail deviates from a straight line. This deviation is likely a consequence of corpus construction, since the dataset was seeded using domain-specific keywords including \textit{biocomposite} and \textit{scaffolds}. These terms appear very frequently across the documents and therefore form disproportionately large hubs, which slightly distort the ideal power law decay. Even with this effect, the overall shape of the distribution reveals a scale-free structure that is typical of scientific corpora. A relatively small set of broadly used concepts forms the dense connective backbone of the hypergraph, while a long tail of specialized or context-dependent terms appears only sparsely throughout the literature. This pattern reflects both the thematic organization of the field and the inherent heterogeneity of the domain.

\begin{figure}[htbp]
    \centering
    \includegraphics[width=1\linewidth]{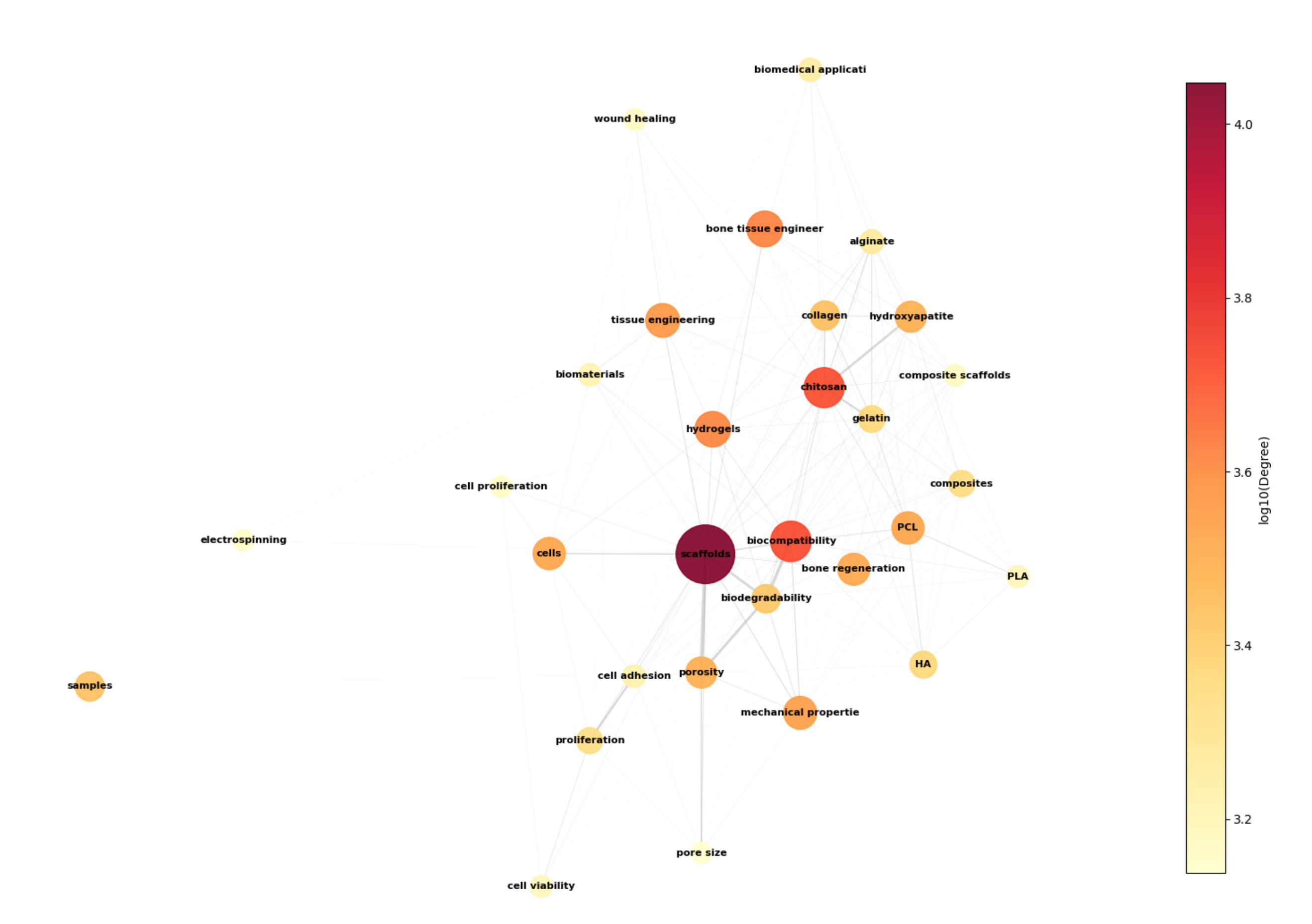}
    \caption{This network visualized of the top 30 highest-degree nodes (“hubs”) in the hypergraph and the strength of their co-occurrence within hyperedges. Each node corresponds to a frequently occurring concept, with node size proportional to its degree (total number of hyperedges it appears in) and node color representing log(degree), where darker red indicates more central, heavily reused concepts. Edges represent co-occurrence relationships between hub pairs, drawn only when two concepts appear together in at least 10 hyperedges; edge thickness increases with co-occurrence frequency, highlighting dominant conceptual pairings. Network Statistics: Nodes: 30, Edges: 207, Density: 0.476, Avg clustering: 0.647.}
    \label{fig:hubs}
\end{figure}

The subnetwork of the top thirty highest-degree nodes provides a focused view of the concepts that structure the corpus at the core. As shown in Figure~\ref{fig:hubs}, these hubs correspond to widely recurring terms such as \textit{scaffolds, biocompatibility, chitosan, bone tissue engineering,} and \textit{hydrogels}, all of which are central topics in biocomposite scaffolding research. The strong concentration of edges among these nodes reflects frequent co-occurrence within hyperedges, indicating that these concepts routinely appear together in the same scientific contexts. The high network density of 0.476 and average clustering coefficient of 0.647 further reveal a tightly interconnected conceptual core, where many hub nodes share multiple overlapping relationships. Several thematic clusters are also evident: one grouping centers around polymeric biomaterials such as \textit{PCL, PLA, and gelatin}; another connects biological processes including \textit{cell adhesion, porosity, and proliferation}; and a third links structural or functional properties relevant to scaffold design. The central placement of scaffolds in particular is consistent with its large global degree and with its role as a primary keyword used during corpus collection. The hub landscape confirms that the corpus is organized around a small set of dominant concepts that anchor the majority of scientific discussions, while secondary topics branch out from these foundational themes. This structure aligns with the power law behavior observed earlier and illustrates how higher-order co-occurrence patterns give rise to a coherent conceptual backbone within the hypergraph.

Table~\ref{tab:hub_ego_analysis} provides a more quantitative view of the hub structure by reporting global degree, ego-network connectivity, and dominant co-occurring concepts for the top twenty nodes in the hypergraph. Several clear patterns emerge from this analysis. First, the degree distribution within the hub set spans a wide range, from over 11{,}100 occurrences for \textit{scaffolds} to approximately 1{,}800 for \textit{alginate}, confirming a strong hierarchy within the conceptual core. Interestingly, although \textit{biocomposite scaffolds} served as the seed term for constructing the corpus, the node \textit{biocomposite} itself does not emerge as a hub. The second largest hub node is \textit{biocompatibility}, which is obviously a key feature of biocomposite scaffolds. Although \textit{scaffolds} accounts for 3.60\% of all hyperedges in the corpus, the next several hubs, including \textit{biocompatibility} (1.72\%) and \textit{chitosan} (1.60\%), maintain considerably lower but still substantial contributions to the global edge set. Second, ego-network statistics reveal distinct differences in local structural roles. Concepts such as \textit{mechanical properties, HA, }and\textit{ gelatin} exhibit neighbor densities above 0.80, indicating that their surrounding neighborhoods are densely interconnected and form compact clusters. This pattern suggests that within the corpus on biocomposite scaffolds these concepts represent strongly co-defined design parameters rather than isolated topics. Their high local density reflects the way scaffold research integrates structural mechanics, pore architecture, and material formulation into a unified framework, where changes in one property are often discussed alongside consequences for others. In contrast, nodes like \textit{bone tissue engineering} and \textit{samples} display lower densities near 0.50 to 0.56, suggesting that these terms connect across a broader range of loosely related concepts rather than forming a single cohesive neighborhood. Without difficulty, we can understand that these hubs function as conceptual bridges within the corpus.

\begin{table}[htbp]
\centering
\caption{Integrated Hub Node Analysis: Ego Network Metrics and Global Degree Contribution}
\label{tab:hub_ego_analysis}
\scriptsize
\begin{tabular}{clrrrrl}
\hline
\textbf{Rank} & \textbf{Concept} & \textbf{Degree} & \textbf{\% of Edges} & 
\textbf{Unique Neighbors} & \textbf{Neighbor Density} & \textbf{Top Co-occurring} \\
\hline
1 & scaffolds & 11157 & 3.60\% & 5440 & 0.487586 & porosity(967), biodegradability(624), cells(288) \\
2 & biocompatibility & 5329 & 1.72\% & 4103 & 0.769938 & biodegradability(799), scaffolds(261), bioactive(225) \\
3 & chitosan & 5211 & 1.60\% & 3697 & 0.709461 & hydroxyapatite(589), gelatin(452), collagen(276) \\
4 & bone tissue engineering & 4148 & 1.29\% & 2153 & 0.519045 & HAp-based composites(213), scaffolds(133), hydroxyapatite(66) \\
5 & hydrogels & 4076 & 1.33\% & 2747 & 0.673945 & biocompatibility(104), tissue engineering(68), porosity(67) \\
6 & tissue engineering & 3713 & 1.15\% & 2383 & 0.641799 & drug delivery(216), regenerative medicin(175), scaffolds(161) \\
7 & mechanical properties & 3521 & 1.12\% & 2980 & 0.84635 & biocompatibility(200), scaffolds(177), porosity(98) \\
8 & PCL & 3369 & 1.02\% & 2651 & 0.786888 & PLA(161), gelatin(134), HA(128) \\
9 & cells & 3360 & 1.07\% & 2068 & 0.615476 & scaffolds(288), hydrogels(65), growth factor(44) \\
10 & bone regeneration & 3317 & 1.05\% & 2233 & 0.673199 & scaffolds(115), hydroxyapatite(51), composite scaffolds(40) \\
11 & porosity & 3154 & 0.98\% & 2019 & 0.64014 & scaffolds(967), biodegradability(651), pore size(300) \\
12 & hydroxyapatite & 3117 & 0.96\% & 2088 & 0.669875 & chitosan(589), chitosan/hydroxyapat(174), collagen(153) \\
13 & collagen & 2788 & 0.87\% & 2079 & 0.745696 & chitosan(276), a structural protein(259), hydroxyapatite(153) \\
14 & samples & 2756 & 0.88\% & 1556 & 0.564586 & SEM(63), PBS(50), gold(46) \\
15 & biodegradability & 2632 & 0.85\% & 1717 & 0.652326 & biocompatibility(799), porosity(651), scaffolds(624) \\
16 & HA & 2338 & 0.70\% & 1978 & 0.846022 & PCL(128), chitosan(56), biocompatibility(49) \\
17 & gelatin & 2303 & 0.74\% & 1915 & 0.831524 & chitosan(452), alginate(157), collagen(139) \\
18 & composites & 2234 & 0.72\% & 1951 & 0.873321 & hydroxyapatite(83), chitosan(51), biocompatibility(46) \\
19 & proliferation & 2182 & 0.68\% & 2126 & 0.974335 & cell adhesion(504), differentiation(372), cell attachment(291) \\
20 & alginate & 1847 & 0.60\% & 1591 & 0.861397 & chitosan(193), gelatin(157), collagen(113) \\
\hline
\end{tabular}
\end{table}

The unique neighbors also highlights variation in conceptual breadth. \textit{Scaffolds} has more than 5{,}400 unique neighbors, far exceeding all other hubs and reflecting its central role in the corpus. \textit{Biocompatibility} and \textit{chitosan} follow with roughly 4{,}100 and 3{,}700 neighbors, respectively, reinforcing their importance as widely reused foundational concepts. In contrast, specialized topics such as \textit{hydroxyapatite}, \textit{PCL}, and \textit{HA} have between 1{,}950 and 2{,}300 neighbors, consistent with their more focused application domains within bone regeneration or composite materials. Finally, the top co-occurring concepts explicitly describe the thematic signatures for each hub. \textit{Scaffolds} is most frequently paired with \textit{porosity}, \textit{biodegradability}, and \textit{cells}, emphasizing structural, functional, and biological design considerations. \textit{Biocompatibility} frequently co-occurs with \textit{scaffolds}, \textit{bioactive materials}, and \textit{porosity}, underscoring its central role in materials safety and biological integration. Polymer-based hubs such as \textit{PCL} and \textit{PLA} are consistently linked to one another, whereas biological hubs like \textit{cells}, \textit{proliferation}, and \textit{cell adhesion} cluster around mechanobiological processes.

To deepen our understanding of how central concepts interact within the corpus, it is useful to move beyond individual degree counts and ego-network densities and examine how hubs relate to one another collectively. While previous analyses revealed which concepts dominate the global structure and how they cluster locally, they do not fully capture the degree to which high-importance concepts form an integrated backbone. For a corpus focused on biocomposite scaffolds, identifying this backbone is particularly important since scientific advances in this domain often depend on the coordinated interplay between material properties, biological responses, and fabrication strategies. A hub-centric analysis therefore provides a natural next step for probing whether the most influential concepts operate independently or whether they form a rich, tightly interconnected core that drives the thematic coherence of the literature.

\begin{table}[htbp]
\centering
\caption{Hub Integration Scores and Rich-Club Connectivity}
\label{tab:richclub}

\begin{tabular}{l r}
\hline
\textbf{Hub Integration Scores (total co-occurrences with other hubs)} & \textbf{Score} \\
\hline
scaffolds & 3335 \\
biodegradability & 2506 \\
chitosan & 2260 \\
biocompatibility & 1948 \\
porosity & 1942 \\
gelatin & 1266 \\
hydroxyapatite & 1258 \\
collagen & 996 \\
alginate & 803 \\
PCL & 801 \\
\hline
\end{tabular}

\vspace{0.5cm}

\begin{tabular}{c c c c}
\hline
\textbf{Rich-Club Degree Threshold} &
\textbf{\# Hub Nodes} &
\textbf{Edges Among Hubs} &
\textbf{Rich-Club Coefficient} \\
\hline
10 & 9295 & 115698 & 0.002679 \\
20 & 4445 & 85438  & 0.00865 \\
50 & 1611 & 53481  & 0.041239 \\
100 & 701 & 34996 & 0.142637 \\
\hline
\end{tabular}

\end{table}

Table~\ref{tab:richclub} summarizes two complementary perspectives on this structure. The first is the hub integration score, which measures how frequently each high-degree concept co-occurs with other hubs. Here, \textit{scaffolds} again dominates with 3{,}335 co-occurrences, reflecting its role as the conceptual anchor of the field. \textit{Biodegradability}, \textit{chitosan}, \textit{biocompatibility}, \textit{porosity}, and \textit{gelatin} also rank highly, with scores between 1{,}266 and 2{,}506. These values indicate that these materials and scaffold properties consistently appear alongside other core concepts, reinforcing their status as central design considerations within biocomposite scaffold research. In contrast, \textit{hydroxyapatite}, \textit{collagen}, \textit{alginate}, and \textit{PCL} have lower integration scores, suggesting that although they are important materials, they participate in more specialized or context-dependent discussions. The second perspective is provided by the rich-club analysis, which examines whether high-degree nodes preferentially connect to one another. Increasing the degree threshold from 10 to 100 reduces the size of the hub set from 9{,}295 to 701 nodes, yet the rich-club coefficient rises sharply from 0.002679 to 0.142637. This pattern indicates that the highest-degree concepts form a progressively more interconnected subnetwork as the threshold increases. In other words, the most influential concepts in the field are not only highly reused but also strongly interlinked.

The extraction of s-connected components as shown in Table \ref{tab:s_components} offers a principled framework for characterizing higher-order organization within the hypergraph. Formally, an s-connected component is defined as a maximal set of hyperedges in which every pair of hyperedges is linked through a chain of intermediate hyperedges such that each adjacent pair shares at least s nodes. This definition extends classical graph connectivity to the multi-entity setting, ensuring that membership is determined not by binary adjacency but by the persistence of multiway overlap across hyperedges.

\begin{table}[htbp]
\centering
\begin{tabular}{lr}
\toprule
\textbf{Property} & \textbf{Value} \\
\midrule
\multicolumn{2}{l}{$s=1$} \\
\quad \# of connected components & 3011 \\
\quad Largest component size & 153{,}987 edges \\
\quad Top 10 component sizes & [153987, 19, 14, 13, 12, 11, 10, 10, 8, 8] \\
\midrule
\multicolumn{2}{l}{$s=2$} \\
\quad \# of connected components & 2618 \\
\quad Largest component size & 27{,}901 edges \\
\quad Top 10 component sizes & [27901, 19, 10, 10, 9, 9, 9, 9, 8, 8] \\
\midrule
\multicolumn{2}{l}{$s=3$} \\
\quad \# of connected components & 1151 \\
\quad Largest component size & 9{,}082 edges \\
\quad Top 10 component sizes & [9082, 15, 13, 9, 7, 7, 6, 6, 6, 6] \\
\midrule
\multicolumn{2}{l}{$s=4$} \\
\quad \# of connected components & 605 \\
\quad Largest component size & 4{,}503 edges \\
\quad Top 10 component sizes & [4503, 30, 11, 8, 7, 6, 6, 6, 6, 6] \\
\bottomrule
\end{tabular}
\caption{$s$-connected components for $s=1$ to $s=4$. Larger $s$ values reveal tightly bound, mature conceptual clusters; smaller $s$ show more diffuse or emerging areas.}
\label{tab:s_components}
\end{table}

As s increases, only those hyperedges that repeatedly co-occur in densely overlapping conceptual contexts remain connected, thereby isolating highly structured regions of the corpus. In contrast to pairwise graph representations, $s$-components preserve the integrity of the multi-concept units that underpin scientific discourse. Components at low $s$ reflect broad and loosely associated topical regions, typically corresponding to heterogeneous or exploratory areas of research. Intermediate values of s reveal subdomains with moderate conceptual cohesion, where material properties, biological processes, and scaffold characteristics are discussed in recurring but flexible combinations. High-$s$ components, by contrast, isolate stable conceptual ecosystems defined by repeated co-occurrence of specific material formulations, design parameters, or mechanobiological relationships across many studies.

Extracting this structure opens several analytically valuable directions. It enables the identification of thematic stability within the corpus, helping to distinguish emerging, exploratory scaffold designs from well-established combinations of polymer chemistry, cell behavior, and mechanical evaluation. The $s$-component framework also provides a principled mechanism for noise reduction, emphasizing hyperedges embedded in repeated higher-order overlaps while downweighting incidental or isolated co-occurrences. The stratified representation naturally supports downstream multi-agent reasoning workflows, with high-$s$ components serving as stable grounding regions for scientific inference, intermediate-s components facilitating cross-domain hypothesis generation, and low-$s$ components revealing areas where conceptual integration remains in flux.

\begin{figure}[htbp]
    \centering
    \includegraphics[width=1.0 \linewidth]{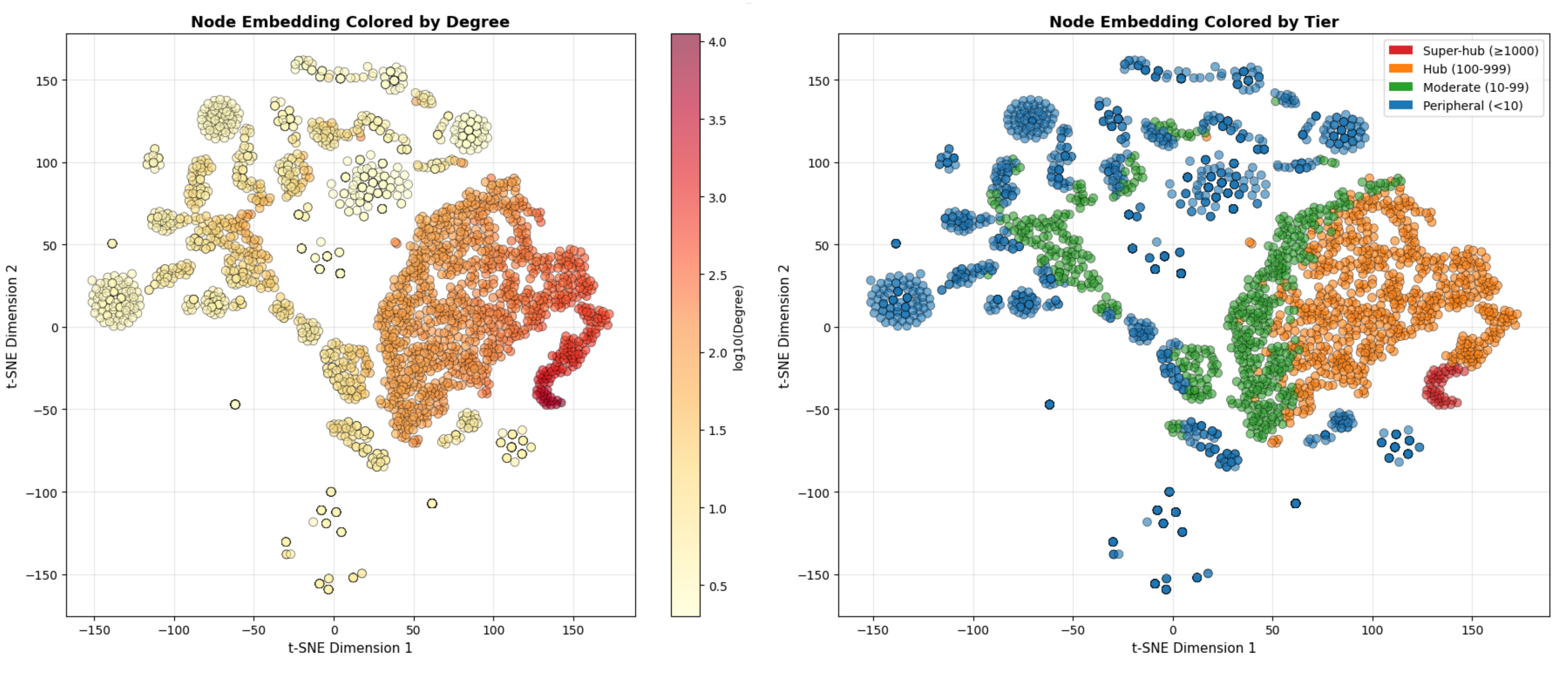}
    \caption{t-SNE projection of nodes based on three-dimensional structural signatures (degree, unique neighbors, average hyperedge size). Left: nodes colored by log-degree show high-degree concepts collapsing into a compact core while low-degree concepts fragment into isolated peripheral islands. Right: tier-based coloring (super-hubs in red, hubs in orange, moderate in green, low in blue) reveals a continuous gradient bridging the dense conceptual nucleus and heterogeneous periphery, indicating that the corpus exhibits layered conceptual roles rather than discrete modules.}
    \label{fig:tsne}
\end{figure}

To complement the global and local analyses described above, we employed t-distributed Stochastic Neighbor Embedding (t-SNE) shown in Figure \ref{fig:tsne} to examine the geometry of node-level structural roles within the hypergraph. Whereas degree distributions, ego-network densities, rich-club connectivity, and s-connected components quantify specific aspects of connectivity, they do not directly reveal how these structural attributes jointly organize concepts in the underlying feature space. To address this, each node was represented by a three-dimensional structural signature comprising its global degree, the number of unique neighbors with which it co-occurs, and the average size of the hyperedges to which it belongs. These features capture how frequently a concept is reused, how broadly it participates across contexts, and whether it tends to appear in small, specific hyperedges or in large, multi-entity conceptual groupings. After standardization, the resulting feature matrix was projected into two dimensions using t-SNE to visualize the relative similarity of structural roles across the corpus.

The embedding exposes a critical distinction not easily visible from univariate metrics alone: connectivity magnitude does not predict structural role. High-degree concepts collapse into a single, dense cluster, revealing that all central terms participate through identical structural patterns (similar neighbor diversity, similar hyperedge sizes) and form a homogeneous, unified conceptual backbone. In contrast, peripheral concepts fragment into numerous discrete islands despite having comparably low degrees, demonstrating that "low-frequency" can encompass multiple fundamentally distinct participation modes: each island represents concepts that share structural signatures within the island but exhibit entirely different neighbor diversity and hyperedge participation patterns between islands. This fragmentation reveals what degree alone obscures: nodes with identical connectivity can occupy non-overlapping structural niches. Moderate-degree concepts form a continuous gradient bridging these regimes evidenced in the right panel of Figure \ref{fig:tsne} by green nodes spatially distributed between the dense orange-red cluster and the fragmented blue islands. This indicates that conceptual consensus develops through gradual standardization rather than discrete phase transitions.

This geometry clarifies field development. The homogeneous core demonstrates that biocomposite scaffold research operates from a structurally unified foundation where central design principles recur with consistent participation patterns across all contexts, forming a single shared vocabulary rather than competing paradigms. This is insightful as it is somewhat rare, as many fields show persistent modularity reflecting competing approaches or subdisciplinary boundaries. The unified core in biocomposite scaffold research indicates that the field has achieved conceptual consolidation. The fragmented periphery reveals that innovation occurs through parallel, structurally isolated exploration: novel materials, fabrication techniques, and assays emerge in tightly coupled specialist niches, each island developing distinct structural signatures that prevent cross-pollination despite operating at the same connectivity scale. 

The multi-dimensional embedding thus exposes that concepts in this field achieve centrality not by increasing connections alone, but by converging toward the homogeneous structural signature of the core where consensus in the field is found. For instance, a biomaterial appearing in numerous publications but used inconsistently across disparate application domains, with one group investigating it for drug delivery using pharmacological characterization, another for tissue scaffolds using cell biology assays, and a third for coatings using surface chemistry methods, would exhibit high degree yet remain structurally peripheral due to divergent neighbor sets and variable hyperedge participation. In contrast, a material achieving structural centrality must be deployed consistently: characterized with standardized methods, paired with compatible processing techniques, and embedded in similar experimental frameworks across studies. Structural convergence thus represents a necessary condition for broad adoption, as reproducible science requires stable, predictable conceptual frameworks that enable researchers to build on prior work, integrate findings across studies, and establish cumulative knowledge rather than fragmented, context-specific observations.

\subsection{Agentic Reasoning on a Hypergraph}\label{sec5}

\begin{figure}
    \centering
    \includegraphics[width=1\linewidth]{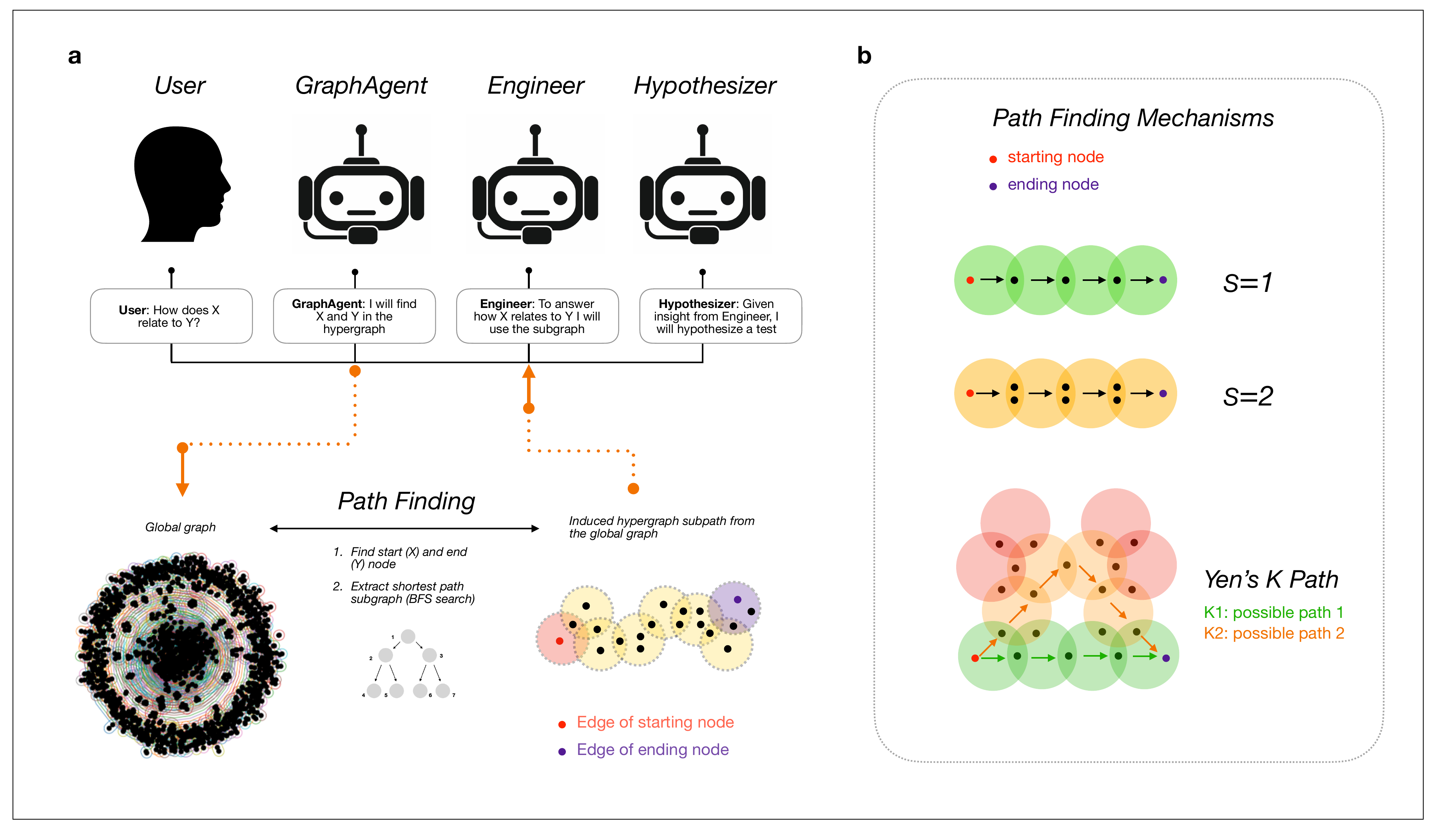}
    \caption{(a) Overview of the multi-agent reasoning system. The User submits a scientific question connecting two concepts (X and Y). The GraphAgent locates these entities in the global hypergraph and extracts an induced subgraph representing the shortest relational structure between them. The Engineer interprets this subgraph mechanistically, and the Hypothesizer proposes testable hypotheses based on the inferred mechanism. (b) Illustration of allowable hypergraph traversal mechanisms. Paths are recovered under a node intersection constraint (S), where adjacent hyperedges must share exactly one (S=1) or two (S=2) nodes. A Yen-style k-shortest path strategy then identifies multiple alternative minimal-length hyperpaths (K1, K2), enabling richer reasoning substrates.}
    \label{fig:agents}
\end{figure}

We next integrate the hypergraph into an agentic framework that leverages its structural topology to support multi-step reasoning. Within this framework, agentic interactions are initiated by a user-provided query. The GraphAgent is responsible for extracting the scientifically relevant keywords, embedding these terms, and aligning them with the closest corresponding nodes in the hypergraph. These matched nodes serve as the anchors for computing shortest hypergraph paths, thereby identifying the sequence of incident hyperedges and intermediate nodes that provide the most parsimonious connection between the start and end nodes. The resulting subgraph representation is then transmitted to the Engineer agent, who synthesizes this structured information to generate an informed response to the original query. Finally, a Hypothesizer agent builds upon the Engineer’s analysis to propose a novel experimental hypothesis. The complete agentic workflow is depicted in Figure~\ref{fig:agents}.

One practical use of the hypergraph is to identify how a low-degree concept might be mechanistically connected to a high-degree hub. By selecting a high-degree node and pairing it with a sparsely connected node, the system can trace the shortest available path between them and use this path to infer potential combinations, composite formulations, or hypotheses for further study. In effect, this approach strengthens the integration of low-degree concepts into the broader knowledge network by leveraging the relational context encoded in the hypergraph edges.

Drawing on our previous analysis of the top 20 hubs, we designate \textit{PCL} as the representative high-degree node and, illustratively, select \textit{cerium oxide} as a low-degree node. The agent conversation is shown in Figure~\ref{fig:cerium}. We impose a minimum hyperedge intersection size of one shared node (IS = 1) and extract only the shortest hypergraph path (K = 1) connecting the two concepts. 

Notably, the graph path identified by the GraphAgent is considerably short but does succesfully find a route between \textit{cerium oxide} and \textit{PCL}. As a consequence, the Engineer and Hypothesizer leverages what it is given as context to propose a minimum viable plan which is to embed cerium oxide nanoparticles within PCL nanofibers as seen in Figure~\ref{fig:cerium}. The protocol developed by the hypothesizer is notably detailed and sensible, outlining the synthesis of cerium oxide nanoparticles using methods such as hydrothermal or sol–gel processing, followed by the preparation of PCL–cerium oxide nanofibers via electrospinning from a composite solution containing both PCL and the nanoparticles. The agent then proposes a comprehensive evaluation pipeline that is reasonable, including materials characterization, performance testing, and cytotoxicity assessment.

\begin{rightbubbles}
\textbf{USER:} How can cerium oxide mechanistically relate to PCL? 
\end{rightbubbles}

\begin{leftbubbles}
\textbf{GRAPHAGENT:} Extracted keywords: \texttt{['cerium oxide', 'pcl']}. 
Found matched nodes in embeddings: \texttt{[np.str\_('Cerium oxide'), np.str\_('PCL')]}.

\textbf{GRAPHAGENT:} Please consider the following relationships of the knowledge related to the question and make your response: 

\textbf{GRAPHAGENT:} \texttt{[['Cerium oxide, Chitosan, Hydroxyethylcellulose, Polyethylene glycol compose Antibacterial nano composite films.', 'PCL, chitosan compose PCL/chitosan nanofibers.']]}.
\end{leftbubbles}

\begin{rightbubbles}
\textbf{ENGINEER:} Cerium oxide may mechanistically relate to Poly(caprolactone) (PCL) through a potential composite material formation, where cerium oxide nanoparticles could be incorporated into PCL-based nanofibers, similar to how chitosan is combined with PCL to form PCL/chitosan nanofibers [1]. This hypothetical relationship suggests that the antibacterial properties of cerium oxide, as observed in its composition with other polymers like chitosan and hydroxyethylcellulose in nano composite films [2], could potentially be leveraged in conjunction with PCL to create novel materials with enhanced properties. However, this proposed connection is speculative and would require further investigation to determine the feasibility and characteristics of such a composite material.
\end{rightbubbles}

\begin{leftbubbles}
\textbf{HYPOTHESIZER: } To further explore the mechanistic relationship between cerium oxide and PCL, an experiment could be designed to synthesize and characterize a composite material consisting of cerium oxide nanoparticles embedded within PCL nanofibers. 

\textbf{HYPOTHESIZER:} This could involve:
\begin{enumerate}
    \item \textbf{Synthesis of Cerium Oxide Nanoparticles}: Utilizing a method such as hydrothermal synthesis or sol-gel processing to produce cerium oxide nanoparticles with controlled size and morphology.
    \item \textbf{Preparation of PCL/Cerium Oxide Composite Nanofibers: }Employing electrospinning or a similar technique to fabricate nanofibers from a solution containing PCL and the synthesized cerium oxide nanoparticles. The concentration of cerium oxide and the processing conditions could be varied to optimize the dispersion of nanoparticles within the PCL matrix.
    \item \textbf{Characterization of Composite Nanofibers:} Using techniques such as transmission electron microscopy (TEM), scanning electron microscopy (SEM), and X-ray diffraction (XRD) to examine the morphology, structure, and distribution of cerium oxide nanoparticles within the PCL nanofibers.
    \item \textbf{Evaluation of Mechanical and Antibacterial Properties:} Conducting tensile tests to assess the mechanical strength and elasticity of the composite nanofibers, and performing antibacterial assays (e.g., against E. coli or S. aureus) to determine if the incorporation of cerium oxide nanoparticles imparts significant antimicrobial activity to the PCL nanofibers.
    \item \textbf{Cytotoxicity and Biocompatibility Assessment:} Evaluating the cytotoxicity of the composite material using cell culture assays (e.g., MTT assay) to ensure its biocompatibility, which is crucial for potential biomedical applications.
By investigating the properties and behavior of cerium oxide-PCL composite nanofibers through such an experiment, researchers could gain insights into how these components interact at a mechanistic level, potentially uncovering new avenues for the development of advanced biomaterials with tailored functionalities.
\end{enumerate}

\textbf{HYPOTHESIZER}: By investigating the properties and behavior of cerium oxide-PCL composite nanofibers through such an experiment, researchers could gain insights into how these components interact at a mechanistic level, potentially uncovering new avenues for the development of advanced biomaterials with tailored functionalities.
\end{leftbubbles}

\begin{figure}[htbp]
\centering
\small
\caption{Agentic dialogue initialized with the query: “How does cerium oxide mechanistically relate to PCL?” The hypergraph was queried using an intersection size of one shared node (IS = 1) and a single shortest hypergraph path (K = 1). The resulting dialogue illustrates how the agent incorporates hypergraph structure to propose a mechanistic bridge that increases connectivity between a sparsely connected entity and a densely connected one.}
\label{fig:cerium}
\end{figure}

For our next experiment, we investigate a mechanistic relationship between PCL and a low-degree node in the graph that, at first glance, appears unrelated to biocomposite scaffolds in a tissue-engineering context. To illustrate this capability, we focus on the node \textit{grass}, selected arbitrarily for exploration. Results are shown in in Figure~\ref{fig:grass}. Remarkably, the graph agent uncovers that fescue grass used in processes that yield hydrogen and biomass-derived methanol, forms a mechanistic link to PCL precipitation. Building on this, the hypothesizer agent proposes leveraging this pathway to produce PCL, outlining an experimental design involving biomass conversion, methanol purification, and subsequent PCL precipitation and characterization. 

%%% GRASS AND PCL 

\begin{rightbubbles}
\textbf{USER:} How can grass mechanistically relate to PCL? 
\end{rightbubbles}

\begin{leftbubbles}
\textbf{GRAPHAGENT:} Extracted keywords: \texttt{['grass', 'pcl']}. 
Found matched nodes in embeddings: \texttt{[np.str\_('fescue grass'), np.str\_('PCL')]}.

\textbf{GRAPHAGENT:} Please consider the following relationships of the knowledge related to the question and make your response: 

\textbf{GRAPHAGENT:} \texttt{['fescue grass achieves hydrogen production rate.', 'sacrifcial electron donors improve hydrogen production rates.', 'formic acid, methanol investigated as sacrifcial electron donors.', 'PDLLA, PCL precipitated into methanol.']}.
\end{leftbubbles}

\begin{rightbubbles}
\textbf{ENGINEER:} Fescue grass has been found to achieve a hydrogen production rate [1], and sacrificial electron donors, such as formic acid and methanol, have been investigated to improve hydrogen production rates [2]. Notably, methanol is also used as a solvent to precipitate polymers like PCL (Polycaprolactone) and PDLLA (Poly(D,L-lactic acid)) [3]. Therefore, it can be inferred that the relationship between grass and PCL is indirect, with grass potentially contributing to hydrogen production, which in turn could be related to the use of methanol as a sacrificial electron donor, and ultimately, as a solvent for precipitating PCL. However, this connection is hypothetical and requires further investigation.
\end{rightbubbles}

\begin{leftbubbles}
\textbf{HYPOTHESIZER: }To further explore the mechanistic relationship between grass and PCL, an experiment could be designed to investigate the potential of using biomass-derived methanol from grass as a solvent for precipitating PCL. This could involve several steps:

\textbf{HYPOTHESIZER:} The hypothesis is that the incorporation of cerium oxide into a PCL-chitosan composite would not only provide antibacterial properties due to the presence of cerium oxide but also enhance the mechanical strength and biocompatibility of the material. The use of chitosan as a natural polymer component could facilitate cell adhesion and proliferation, while PCL would contribute to the structural integrity and durability of the scaffold.

\textbf{HYPOTHESIZER:} The experimental design could involve the following steps:
\begin{enumerate}
    \item \textbf{Biomass Conversion:} Convert fescue grass or other types of grass into biomass-derived methanol through a process such as gasification or biochemical conversion.
    \item \textbf{Purification:} Purify the biomass-derived methanol to ensure it meets the required standards for use as a solvent in polymer precipitation.
    \item \textbf{PCL Precipitation:} Use the purified biomass-derived methanol as a solvent to precipitate PCL, comparing the results with traditional methods using commercially available methanol.
    \item \textbf{Characterization:} Characterize the physical, chemical, and mechanical properties of the PCL precipitated using biomass-derived methanol to determine if there are any differences or advantages compared to traditionally precipitated PCL.
\end{enumerate}

\textbf{HYPOTHESIZER:} This experiment could reveal new insights into the potential for using renewable biomass sources like grass to produce solvents for polymer precipitation, potentially offering a more sustainable alternative to traditional methods. Additionally, it could explore the properties of PCL precipitated from biomass-derived methanol, which might exhibit unique characteristics due to the solvent's origin. Furthermore, considering the role of sacrificial electron donors in improving hydrogen production rates, another area of investigation could involve using formic acid or other compounds derived from grass or biomass as sacrificial electron donors to enhance hydrogen production. This could lead to a more integrated system where grass contributes to both hydrogen production (through its biomass) and the precipitation of polymers like PCL (through the use of biomass-derived solvents).
\end{leftbubbles}

\begin{figure}[htbp]
\centering
\small
\caption{Agentic dialogue with the query: “How does grass mechanistically relate to PCL?” The hypergraph was queried with one shared intersection node (IS = 1) and a single shortest hypergraph path (K = 1), illustrating how hypergraph-aware reasoning proposes a mechanistic link between semantically distant entities.}
\label{fig:grass}
\end{figure}

%% k = 3 

To provide a larger substrate for hypothesis generation, we extend the GraphAgent to traverse the hypergraph and return the top 3 shortest hypergraph paths (K = 3) between nodes using a Yen-style K-shortest paths procedure. The traversal enforces a node intersection size of one (IS = 1), ensuring that each step shares exactly one common node. As seen in Figure~\ref{fig:Yen3}, we revisit the relation between \textit{cerium oxide} and \textit{PCL} where the GraphAgent produces these 3 hyperedge paths. From this context, the Engineer agent identifies the overlapping node \textit{chitosan} as the mechanistic bridge between the \textit{PCL} and \textit{cerium-oxide} associated hyperedges. Because each component appears in composite formulations within the corpus, the path suggests that a new composite involving all three may be plausible. The agent even hypothesizes that the resulting material would be fibrous, drawing on the known hypergraph relationship between PCL and chitosan, which are documented to form nanofibrous structures. The hypothesis agent outlines an experimental workflow involving the synthesis of cerium oxide nanoparticles, independent fabrication of PCL–chitosan nanofibers, and their subsequent integration into a reinforced composite scaffold. 

Compared with the earlier agentic hypothesis, which relied on a single hypergraph path, the absence of additional contextual paths led to a de-emphasis of chitosan and thus chitosan was omitted from the proposed composite. However, here with additional context we see that the Hypothesis agent recognizes a more robust hypothesis by leveraging chitosan as as a central constituent of the composite. Furthermore, in this iteration the agent not only proposes a new composite material but also specifies its use as a scaffold, identifying applications in biomedical devices and tissue engineering.

\begin{rightbubbles}
\textbf{USER:} How can cerium oxide mechanistically relate to PCL? 
\end{rightbubbles}

\begin{leftbubbles}
\textbf{GRAPHAGENT:} Extracted keywords: \texttt{['cerium oxide', 'pcl']}. 
Found matched nodes in embeddings: \texttt{[np.str\_('Cerium oxide'), np.str\_('PCL')]}.

\textbf{GRAPHAGENT:} Please consider the following relationships of the knowledge related to the question and make your response: 

\textbf{GRAPHAGENT:} \texttt{['Cerium oxide, Chitosan, Hydroxyethylcellulose, Polyethylene glycol compose Antibacterial nano composite films.', 'PCL, chitosan compose PCL/chitosan nanofibers.', 'Cerium oxide, Chitosan, Hydroxyethylcellulose, Polyethylene glycol compose Antibacterial nano composite films.', 'PCL, chitosan compose hybrid vascular grafts.', 'Cerium oxide, Chitosan, Hydroxyethylcellulose, Polyethylene glycol compose Antibacterial nano composite films.', 'PCL, natural polymers compose chitosan, silk, gelatin.']}.
\end{leftbubbles}

\begin{rightbubbles}
\textbf{ENGINEER:} Cerium oxide can mechanistically relate to PCL through the common component of chitosan, as it has been reported that 'PCL, chitosan compose PCL/chitosan nanofibers' [1] and 'Cerium oxide, Chitosan, Hydroxyethylcellulose, Polyethylene glycol compose Antibacterial nano composite films' [2]. This suggests a potential pathway for the integration of cerium oxide with PCL through chitosan-based composites. Furthermore, it is also noted that 'PCL, natural polymers compose chitosan, silk, gelatin' [3], which implies that PCL can be combined with various natural polymers, including chitosan, to form hybrid materials. Therefore, the relationship between cerium oxide and PCL may be established through the use of chitosan as a bridging component in composite materials.
\end{rightbubbles}

\begin{leftbubbles}
\textbf{HYPOTHESIZER: } To further explore the mechanistic relationship between cerium oxide and PCL, a potential experiment could involve designing a composite material that incorporates both cerium oxide and PCL, using chitosan as a bridging component. This could be achieved by developing a hybrid nanofiber scaffold composed of PCL, chitosan, and cerium oxide nanoparticles.

\textbf{HYPOTHESIZER:} The hypothesis is that the incorporation of cerium oxide into a PCL-chitosan composite would not only provide antibacterial properties due to the presence of cerium oxide but also enhance the mechanical strength and biocompatibility of the material. The use of chitosan as a natural polymer component could facilitate cell adhesion and proliferation, while PCL would contribute to the structural integrity and durability of the scaffold.

\textbf{HYPOTHESIZER:} The experimental design could involve the following steps:
\begin{enumerate}
    \item Synthesis of cerium oxide nanoparticles and their characterization in terms of size, shape, and surface chemistry.
    \item Preparation of PCL-chitosan nanofibers using electrospinning or other suitable techniques.
    \item Incorporation of cerium oxide nanoparticles into the PCL-chitosan nanofibers to form a hybrid composite material.
    \item Evaluation of the mechanical properties, biocompatibility, and antibacterial activity of the resulting composite material.
\end{enumerate}

\textbf{HYPOTHESIZER:} This experiment would provide new insights into the potential applications of cerium oxide-PCL composites in biomedical fields, such as tissue engineering, wound healing, and implantable devices. The results could also shed light on the mechanistic relationships between cerium oxide, PCL, and chitosan, paving the way for the development of novel composite materials with enhanced properties.
\end{leftbubbles}

\begin{figure}[htbp]
\centering
\small
\caption{Agentic dialogue with the query: “How does cerium oxide mechanistically relate to PCL?” For this demonstration, the hypergraph intersection size was constrained to a single shared node (IS = 1), and the shortest-path extraction was expanded to three hypergraph paths (K = 3). Agents leverage the additional hypergraph paths to provide richer contextual grounding, while simultaneously improving its ability to reason about mechanistic relationships suggested by the underlying graph structure.}
\label{fig:Yen3}
\end{figure}

To investigate stronger graph paths in which multiple entities are shared across adjacent hyperedges, we set the minimum intersection size to two nodes (IS = 2) seen in Figure~\ref{fig:hydrogel}. For comparability, we restrict the system to return only a single shortest path (K = 1). To ensure that such an intersecting path exists, we select two high-degree nodes, hydrogel and PCL, as the start and end nodes. The GraphAgent identifies a path whose intermediate intersections include chitosan and collagen, which are not part of the original query. Without explicit prompting, the Engineer agent interprets these intersecting nodes as the mechanistic link between hydrogel and PCL, which indicates that the agent is using the hypergraph to inform its reasoning. 

Building on the Engineer's insight, the Hypothesizer proposes a composite scaffold that incorporates all four components. It suggests that creating this composite will require developing a fabrication strategy in which PCL nanofibers are produced through electrospinning and then encapsulated in a chitosan–collagen hydrogel matrix. The Hypothesizer also outlines characterization approaches to evaluate the performance of the composite structure and identifies important limitations, including the different degradation rates of PCL, chitosan, and collagen. It emphasizes the need to assess the composite under physiological conditions where these degradation profiles become critical.

Requiring a minimum intersection size forces the system to retain only those hyperedge transitions that share substantive multi-entity overlap. Larger intersections such as two or more shared nodes reveal stable multi-component motifs that repeatedly co-occur across different contexts providing agents with denser, higher-quality relational information.

\begin{rightbubbles}
\textbf{USER:} How can hydrogel mechanistically relate to PCL? 
\end{rightbubbles}

\begin{leftbubbles}
\textbf{GRAPHAGENT:} Extracted keywords: \texttt{['hydrogel', 'pcl']}. 
Found matched nodes in embeddings: \texttt{[np.str\_('hydrogel'), np.str\_('PCL')]}.

\textbf{GRAPHAGENT:} Please consider the following relationships of the knowledge related to the question and make your response: 

\textbf{GRAPHAGENT:} \texttt{'chitosan, collagen compose hydrogel.', 'PCL, chitosan, collagen, gelatin form scaffolds.']}.
\end{leftbubbles}

\begin{rightbubbles}
\textbf{ENGINEER:} Hydrogel can mechanistically relate to PCL through the common components used in forming scaffolds, such as chitosan and collagen [1]. Since chitosan and collagen are known to compose hydrogels [2], and these same biomaterials, along with PCL, can form scaffolds [3], it suggests a potential intersection or compatibility between hydrogel matrices and PCL-based structures. This relationship could be exploited in tissue engineering applications where the combination of mechanical strength provided by PCL and the bioactive properties of hydrogels composed of chitosan and collagen could enhance scaffold performance [4]. 
\end{rightbubbles}

\begin{leftbubbles}
\textbf{HYPOTHESIZER: } To further explore the mechanistic relationship between hydrogels and PCL, an experiment could be designed to create a composite scaffold that integrates the strengths of both materials. Here's a potential approach:

\textbf{HYPOTHESIZER:}
\begin{enumerate}
    \item \textbf{Fabrication of Composite Scaffolds}: Develop a method to combine PCL with a hydrogel composed of chitosan and collagen. This could involve electrospinning PCL fibers and then encapsulating them within a hydrogel matrix, or vice versa, where hydrogel microspheres are embedded within a PCL scaffold.
    \item \textbf{Mechanical and Biological Characterization}: Evaluate the mechanical properties (e.g., tensile strength, elasticity) of the composite scaffolds in comparison to pure PCL scaffolds and hydrogels. Additionally, assess their biological performance by culturing cells relevant to tissue engineering applications (e.g., osteoblasts for bone tissue engineering) on these scaffolds.
    \item \textbf{Degradation Studies}: Since both PCL and the components of hydrogels (chitosan and collagen) have different degradation rates, it's crucial to study how the composite scaffold degrades over time under physiological conditions. This would provide insights into its potential longevity and biocompatibility in vivo.
    \item \textbf{In Vivo Studies}: For a more comprehensive understanding, conduct animal studies where these composite scaffolds are implanted to assess their integration with host tissue, their ability to support tissue regeneration, and any adverse reactions.
\end{enumerate}

\textbf{HYPOTHESIZER:} By exploring the intersection of hydrogels and PCL through such experiments, researchers could uncover new insights into creating advanced biomaterials for tissue engineering that combine the beneficial properties of both materials, potentially leading to improved outcomes in regenerative medicine.
\end{leftbubbles}

\begin{figure}[htbp]
\centering
\small
\caption{Agentic dialogue with the query: “How does hydrogel mechanistically relate to PCL?” Using larger intersection size (IS = 2) and one shortest graph path (K = 1), the agent exploits higher-order hypergraph structure to derive a mechanistic link across distinct hyperedges. \textit{Chitosan} and \textit{collagen} were the two intersection nodes bridging \textit{hydrogel} and \textit{PCL} used in the hypothesis as the mechanistic link. The terminal nodes were chosen as high-degree nodes to maximize the likelihood of multi-node intersections.
}
\label{fig:hydrogel}
\end{figure}

\section{Conclusion}\label{sec5}

As LLM capabilities expand, so does the need to provide them with richer informational context and to ensure the reliability of their inferences. This necessitates knowledge representations that are both efficient to query and amenable to continual refinement. Raw text is inadequate for this purpose, as it is computationally inefficient and fails to expose the structural patterns and relationships within a corpus.

This paper introduces a mechanism for constructing hypergraph representations from large scientific datasets and analyzing their essential structural characteristics. Our approaches encodes information as a hypergraph rather than a standard pairwise graph, and we show that higher-order relationships among multiple entities are preserved more faithfully, enabling the capture of multiway interactions that are often important for scientific discourse.

Because composite materials research inherently concerns the interplay of multiple interacting components, we selected a corpus on biocomposite scaffolds to rigorously evaluate both the representational fidelity and the reasoning capabilities afforded by a hypergraph-based knowledge structure. This representation allows us to examine the field’s organizational dynamics, including how concepts cluster, evolve, and relate to one another. In turn, these patterns illuminate the historical development of the domain in biocomposite scaffold design and highlight emerging areas of inquiry. Notably, the resulting hypergraph exhibits scale-free topology akin to results shown in earlier work~\cite{Buehler2024, Ghafarollahi2024}, and leveraging this structure allows us to trace meaningful relationships between high-degree and low-degree concepts, thereby uncovering latent connections that span the broader scientific landscape.

To operationalize this knowledge structure for discovery, we integrate the hypergraph into a multi-agent framework equipped with hypergraph-aware traversal and analytical tools. Within this system, we observed that agents not only benefit from access to multiple subgraph paths, thereby leveraging a larger volume of contextual information, but also make deliberate use of the hypergraph as an evidentiary substrate for inference. In particular, they exploit multiway node intersections to tune their belief system and provide a stronger mechanistic foundation for the hypotheses they construct. These intersections form the backbone of coherent causal chains that ground the agents’ reasoning in the scientific literature, thereby justifying the value of supplying agents with hypergraph-based structures rather than relying solely on unstructured text. Ultimately, this establishes a teacherless framework where agents must resolve topological constraints to validate their reasoning, ensuring that discovery is driven by structural necessity rather than statistical imitation \cite{buehler2025selectiveimperfectiongenerativeframework}.

\section{Methods and Materials}\label{sec6}
\subsection{Ontological Hypergraph Corpus Construction}

The initial corpus for hypergraph construction was gathered through searches in the Web of Science Core Collection \cite{clarivate2025} using the query \emph{“biocomposite scaffold.”} Articles that could not be retrieved were predominantly excluded due to incomplete or inaccurate metadata or non-English text. Full-text manuscripts were obtained through a combination of publisher-provided APIs and manual, permission-based scraping.The initial dataset consisted of 1,297 papers collected on July 10, 2025. After cleaning, the final corpus contained 1,097 papers.

\subsection{Manuscript to Hypergraph Algorithm} 
The retrieved PDFs are first converted to Markdown (\texttt{.md}) using \texttt{marker-pdf} (\url{https://github.com/datalab-to/marker}), which preserves core text structure, including section headers, the positions of tables and figures, and in-text reference labels, therefore facilitating downstream LLM-based knowledge extraction. After distillation, each multi-entity extraction is converted into a structured graph fragment encoded using two Pydantic models, Event and Hypergraph, which enforce schema consistency and type safety throughout the pipeline. An Event represents a single directed multi-entity relation. It generalizes the traditional binary triple format to allow sets of entities on both the source and target side. The source is essentially a list of one or more entity names that jointly act as the origin or input of a relation. The target is a list of entity names representing the outcome, product, or affected components of the event. The relation is a textual descriptor naming the interaction. 
The pythonic scheme uses: 

\begin{verbatim}
    class Event(BaseModel):
        source: List[str]
        target: List[str]
        relation: str 
    
    class Hypergraph(BaseModel):
        events: List[Event]
\end{verbatim}

All graph fragments generated from individual document chunks are subsequently merged into a unified global hypergraph. During this merging process, the synchronized node-merging operation ensures that identical or semantically equivalent entities are consolidated into a single canonical node. This alignment step allows hyperedges originating from different parts of the text to connect consistently, enabling the fragmented local extractions to assemble into a coherent, interconnected global relational structure.

\subsection{Models and Libraries}

The \texttt{meta-llama/Llama-3.3-70B-Instruct} model \cite{llama3modelcard} served as the core language model for the multi-agent system. To reduce memory overhead without sacrificing output quality, we used a Q4 quantized version of the model. The model enables a large context window, which we set to 40,000 tokens during inference collection. We hosted the model locally using llama.cpp \cite{gerganov2023llamacpp} deployed with OpenAI-style API interface. The deployment was configured for full GPU use, with tensors distributed across multiple cards. The \texttt{meta-llama/Llama-4-Maverick-17B-128E-Instruct-FP8} model \cite{llama3modelcard} was used to generate the underlying hypergraphs and was accessed via the Together AI API \cite{togetherai}. 

A set of open-source Python tools supported the graph-construction and retrieval system. Text embeddings were generated using the sentence-transformers library \cite{reimers-2019-sentence-bert} with the \texttt{nomic-ai/nomic-embed-text-v1.5} model \cite{nussbaum2024nomic}. This open-source model was selected for its strong retrieval performance, long-context support, and flexible Matryoshka-style embedding sizes, which enable efficient large-scale semantic search. 

The \texttt{HypernetX} package \cite{praggastis2023hypernetxpythonpackagemodeling} and its parent, the \texttt{NetworkX} package \cite{SciPyProceedings_11} was used to analyze the knowledge graph data structures and keep store of node and edge attributes. The open-source \texttt{AutoGen} framework \cite{wu2023autogenenablingnextgenllm} version 0.2.40 from Microsoft Research was employed to construct the multi-agent architecture. AutoGen provides a programmable communication layer that enables LLM-driven agents to exchange structured messages, invoke external tools, and maintain shared state. Within this framework, agents were configured with persistent memory modules and tool-use capabilities, allowing seamless integration of our hypergraph store, vector retrieval components, and domain-specific Python functions into a unified reasoning workflow.

\subsection{Graph Traversal Tools}

The GraphAgent begins by extracting keywords from the query and mapping them to the closest nodes in the hypergraph’s embedding space. In our experiments, we instruct the agent to identify only concrete scientific entities present in the text such as materials, chemicals, biological entities, properties, although this behavior can be adapted to emphasize specific node types or to impose intermediate waypoints for traversal. These extracted terms are then embedded using the same \emph{nomic} sentence embedding model used to embed the hypergraph, and each keyword is matched to existing nodes within a cosine similarity threshold of 1.5.

The traversal algorithm begins by constructing an inverted index that maps each node to the hyperedges in which it appears. This data structure enables efficient local exploration of the hypergraph, avoiding the need to construct the full $s$-line graph, which would be substantially more expensive to compute. For every unordered pair of query nodes, the algorithm performs a Breadth First Search (BFS) directly over hyperedges, allowing a transition from one hyperedge to another only when they share at least $S$ nodes, as specified by the intersection-size parameter $IS$. By retaining all parent hyperedges that reach a given hyperedge at the same BFS depth, the algorithm is able to reconstruct all equally short paths, up to a user-defined maximum of $K$. For each recovered path, detailed metadata are recorded, including the specific intersection nodes and the full membership of each hyperedge. All hyperedges appearing in any of these shortest paths are then aggregated to form the induced sub-hypergraph used in subsequent reasoning steps.

Finally, the extracted hyperedge sequence is translated into natural-language statements by consulting a metadata dataframe containing the source--relation--target triples for each hyperedge. This allows reconstruction of directional sentences of the form ``source -- relation -- target.’’ GraphAgent forwards the reconstructed statements and the original query to the Engineering agents, supplying the information required to infer a mechanistic explanation that connects the extracted entities.

\section*{Author Contributions}
M.J.B. supervised and directed the research. I.A.S. developed the methodology and performed the experiments. Both authors analyzed the data and wrote the manuscript.

\section*{Funding}
I.A.S. acknowledges that this material is based upon work supported by the U.S. Department of Energy, Office of Science, Office of Advanced Scientific Computing Research, under Award Number DE-SC0026073. M.J.B. and I.A.S. acknowledge funding from the MIT Generative AI Initiative and the MIT Generative AI Impact Consortium (MGAIC). 

\section*{Code and Data Availability}
All code, analysis notebooks, and computational protocols developed in this study are publicly available at \url{https://github.com/lamm-mit/HyperGraphReasoning}. The repository also contains supplementary materials, including the full agent outputs referenced throughout this manuscript. Additional datasets are available at \url{https://huggingface.co/datasets/lamm-mit/HyperGraphReasoning}.

\section*{Supplementary Materials}

The paper includes a list of papers used to construct the dataset as Supplementary Materials (\texttt{Final Composite Paper List.xlsx}). The file includes bibliographic details of the papers including title, authors, and DOI access information. The file is also available at \url{https://huggingface.co/datasets/lamm-mit/HyperGraphReasoning/resolve/main/Final%20Composite%20Paper%20List.xlsx}.

\section*{Competing Interests}

The authors declare no competing interests.

%Bibliography
\bibliographystyle{naturemag}
\bibliography{references, hypergraph_references}

%\printbibliography

\end{document}